\PassOptionsToPackage{backref=page}{hyperref}
\documentclass[11pt]{article}

\newif\ifauthordecided
\authordecidedtrue

\newif\ifarxiv
\arxivtrue
\newif\ifcompact
\compacttrue % make it 'true' or 'false' if you want to keep the public or anonymized version

\newif\ifperfect
\perfectfalse % whether we want to be picky and improve every minor detail

\ifarxiv
    \usepackage{acl}
\else 
    \usepackage[review]{acl}
\fi

\usepackage{times}
\usepackage{latexsym}

\usepackage{tabularx}

\usepackage[T1]{fontenc}
\usepackage[utf8]{inputenc}

\usepackage{microtype}
\usepackage{sec/package}

\usepackage{comment}

\newcommand\ours{\textsc{FactTrack}}

\newcommand\world{world state}
\newcommand\interval{{validity interval}}

\newcommand\baselineOutline{\textsc{Full Outline Detection}}
\newcommand\baselinePairwise{\textsc{Pairwise Detection}}

\newcommand\modelGPTF{GPT-4}
\newcommand\modelGPTT{{GPT-3.5-Turbo}}
\newcommand\modelLLaMA{{LLaMA2-7B-Chat}}

\newcommand\metricPairwise{\textsc{Pairwise Score}}
\newcommand\metricContext{\textsc{Context Score}}

\newcommand\prefact{{pre-fact}}
\newcommand\postfact{{post-fact}}

\newcommand\prefacts{{pre-facts}}
\newcommand\postfacts{{post-facts}}
\newcommand\statfacts{{static facts}}

\title{\ours{}: Time-Aware World State Tracking in Story Outlines}

\author{
{\bf Zhiheng Lyu}$^{1}$\ \ \ \ 
  {\bf Kevin Yang}$^2$\ \ \ \ 
  {\bf Lingpeng Kong}$^1$\ \ \ \ 
  {\bf Daniel Klein}$^2$\\
  $^1$The University of Hong Kong, $^2$UC Berkeley \\
  \texttt{\{zhlyu,lpk\}@cs.hku.hk, \{yangk,klein\}@berkeley.edu}
}

\begin{document}

\maketitle
\begin{abstract}
While accurately detecting and correcting factual contradictions in language model outputs has become increasingly important as their capabilities improve, doing so is highly challenging. 
We propose a novel method, \ours{}, for tracking atomic facts and addressing factual contradictions. Crucially, \ours{} also maintains time-aware validity intervals for each fact, allowing for change over time. 
At a high level, \ours{} consists of a four-step pipeline to update a \world{} data structure for each new event: (1) decompose the event into directional atomic facts; (2) determine the \interval{} of each atomic fact using the \world{}; (3) detect contradictions with existing facts in the \world{}; and finally (4) add new facts to the \world{} and update existing atomic facts.
When we apply \ours{} to contradiction detection on structured story outlines, we find that \ours{} using LLaMA2-7B-Chat substantially outperforms a fair baseline using LLaMA2-7B-Chat, and achieves performance comparable to a GPT4 baseline. Moreover, when using GPT4, \ours{} significantly outperforms the GPT4 baseline.
% Beyond the story outline, we also encourage the field to model time-sensitive facts as opposed to facts that are considered perpetually true or false.%, while also reduces the computation cost.
% Beyond the story outline, we hope our work can also inspire further exploration in the field to model time-sensitive facts as opposed to facts that are considered perpetually true or false.
%, while also reduces the computation cost.
% \kevin{Normally you'd give numbers for how much you outperform. but is it accurate to say you outperform GPT4?
% You could just say you are comparable to GPT4 zero shot while only using LLaMA7B, which is just as good.
% Also, can we remove the claim about reducing complexity? Our method is definitely a little complicated.}\zhiheng{Maybe I mean time complexity?}
\footnote{Our code and data 
\ifarxiv
are at \url{https://github.com/cogito233/fact-track}.
 \else
 have been uploaded to the submission system, and will be open-sourced upon acceptance.
\fi
}

% \kevin{test}

% \zhiheng{test}

% \dk{test}

% \lpk{test}

\end{abstract}
\section{Introduction}
\begin{figure}[th]
    \centering
    \hspace*{-0.8cm} % 向左移动图片，可以根据需要调整数值
    \includegraphics[width=\columnwidth]{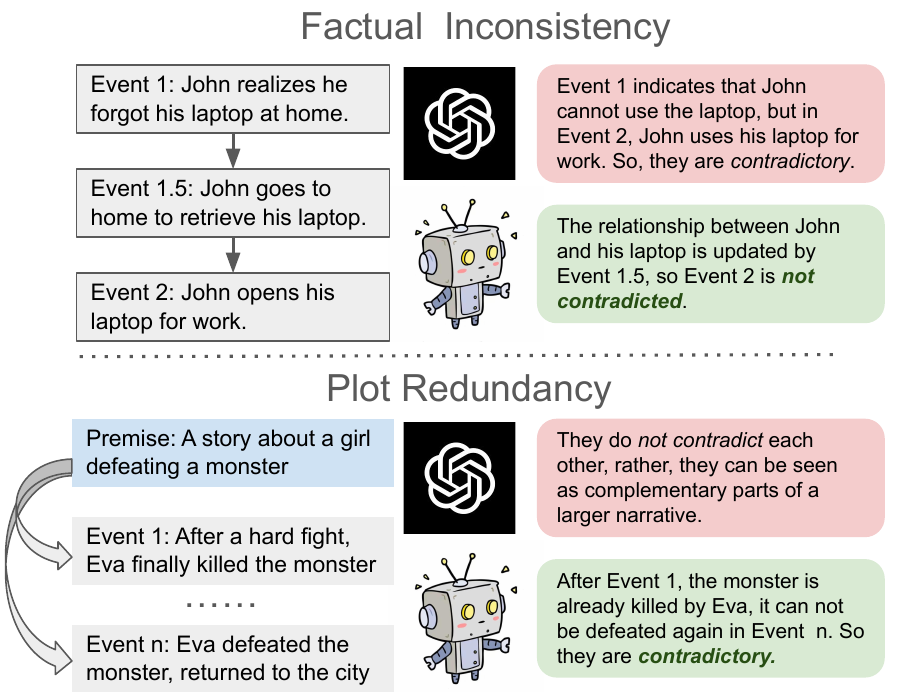} % 调整图片宽度以适应一栏
    \caption{\ours{} tackles the problems of factual inconsistency and plot redundancy. Note that those problems are based on our observations, to provide a clearer understanding of the problem, with both issues considered together in our pipeline and evaluation. For factual inconsistency detection, \ours{} tracks a \interval{} for each fact to distinguish legitimate contradictions from facts simply changing over time. For plot redundancy detection, our method can represent the timeline in more structured form, making detection easier. % \zhiheng{The layout of figure in COLM template seems little bit awkward...}
    %\kevin{use complete sentences (end with a period).
%event 1.5: "john back to home to retrieve his laptop" -> john goes home to retrieve his laptop.
% "So, they are contradicted" -> So, they are contradictory (this occurs in two places.)
% "After event 1, Monster is already killed..." -> "After event 1, the monster is already killed by Eva, so it cannot be defeated again in Event n..."
% basically, proofread your figures too}
    }%\lpk{i would probably say plot 1.5 in the first case, i was reading the left side and i feel it's okay until i saw the red/green explaination. consider plot1 -> plot2; plot 1.5 as a complement}}
    \vspace{-1em}
    \label{fig:analyze}
    % Maybe a figure with Real Examlpes that why our method work and GPT4 not
\end{figure}
Large language models (LLMs) have recently surpassed human performance across a wide range of tasks \citep{ouyang2022training, openai2023gpt4}, yet generating long-form text remains fraught with challenges compared to tasks with shorter outputs.  Even when models are trained to support context windows of hundreds of thousands of tokens, they may still struggle to retrieve and reason over such long context \citep{liu2023lost}. The most advanced existing language models still take long context generation as a direction for further improvement in the future.

Existing works using LLMs in hierarchical generation have explored structured approaches to maintain strong internal coherence within extensive texts of thousands of words \citep{yang2022doc,yang2022re3}. However, challenges remain in maintaining factual consistency and avoiding hallucinations during the generation. Especially when these problems occur in a high-level planning stage, they can greatly damage performance on downstream tasks. Therefore, a system capable of detecting factual contradictions and correcting them is essential. In Figure \ref{fig:analyze}, we take story planning as an example, depicting two common issues encountered: \textit{factual inconsistency} and \textit{plot redundancy}.

% These issues arise partly because large language models are trained on plain text, lack a temporal intuition to align narrative texts with real narrative flows; and partly because, without explicit tracking, these models are insensitive to state updates implied by long texts.\lpk{both of these arguments are a bit unclear to me. First, it's not clear what plain text mean here -- LLMs are trained on resources like BOOKS Corpus, which should contain knowledge required here. Second, it's not sure what is the `temporal intuition' and `state updates implied by long texts', probably say the training sequence usually limited to a certain length and packed with text from random sources sth.} 

Such issues are expected across various domains, as training corpus documents often suffer from length limitations and originate from fragmented sources, impairing the model's ability to establish long-distance, multi-state connections. Moreover, generalizing current sentence-level fact verification models \citep{de-marneffe-etal-2008-finding, cai2021narrative} to accommodate longer texts remains a challenge. We highlight two main distinctions that arise when comparing sentence-level fact statements to lengthier texts: the complexity of multi-fact contexts, and the dynamic, stateful nature of content over time.

To tackle these challenges, Section \ref{sec:atomic_facts} introduces the concept of directional atomic facts, establishing a universal framework for time-aware contradiction detection and state tracking. 
Section \ref{sec:narration_analyze} details the analysis of structures to identify atomic facts and detect contradictions around key events, employing LLMs for event decomposition into \textit{pre-facts} and \textit{post-facts}. 
The concept of pre-facts and post-facts is akin to preconditions and postconditions but formulated in natural text, and implying truthfulness for the entire validity time interval. This approach facilitates tracking the state changes over time prompted by events. 
Maintaining a comprehensive non-contradictory set of facts (\textit{world state}) also allows for detection of contradictions by verifying and updating the atomic facts within the world state. 
Section \ref{sec:pseudo_timeline} explores the broader implications of atomic facts through a more flexible and time-aware framework, introducing a \textit{timeline} for events and atomic facts to identify contradictions or updates through their temporal overlaps.

In Section \ref{sec:maintain_data_structure}, we describe the main implementation of our method, \textbf{\ours{}}. As shown in Figure \ref{fig:pipeline}, we first maintain a list of \prefact{}s and a list of \postfact{}s as our data structure. To update our data structure for a new event, we use a four-step pipeline: Decompose-Determine-Contradiction-Update. With this data structure, we can detect whether two events contradict each other and identify the specific pair of atomic facts in conflict, guiding the subsequent correction procedure. Furthermore, the \interval{} of facts may be useful structural information in and of itself for downstream tasks.

%In this paper, we interpret the Story Plot as forward and backward atomic facts (\prefact{} and \postfact{}) and decompose them using large language models. After obtaining these facts, we maintain a \world, which is the largest independent set of non-contradict facts, for tracking states. (See Section \ref{sec:narration_analyze}) Through this method, we hope to establish a sense of time \lpk{we refers to model? the word 'sense' is also vague here. Avoid using anything that metaphysics in writing :p be as clear as possible what you mean} and ensure consistency when large language models generate narrative texts.

% Furthermore, we extend the above method of chronological order Narrative Fact Tracking and Verification, to the context of Hierarchical Generation through maintaining a fact \interval{} on a pseudo timeline. This extension is meaningful\lpk{avoid judge your own thing, leave it for the readers to judge if it is reasonable or good.}, as it provides a more flexible guideline for Story Planning by allowing Plot elements to be added out of chronological order and checking for contradict with the world state, facilitating the early discovery and resolution of problems at a High-Level plan. In Section \ref{sec:pseudo_timeline}, we discuss how to understand the Pseudo Timeline, the valid intetvals and the conditions of atomic facts; and in Section \ref{sec:maintain_data_structure}, we introduce the details of how we implemented this Data Structure.

% \zhiheng{1.Mention the name; 2.Describe More about Section 4;}

To measure the effectiveness of our approach, we introduce a task for detecting contradictions in the planning procedure, with story outlines serving as our test domain. For event pairs flagged by a method as contradictory, we use \modelGPTF{} to annotate whether they are actually contradictory, on a 1 to 5 scale. Experimental results show that contradictions flagged by \ours{} (using \modelLLaMA{} as a base LLM) are judged to be real contradictions by a score margin of 0.384 more than a fair baseline using \modelLLaMA{}. Additionally, when \ours{} is run on GPT4, the performance significantly surpasses all baselines, including the one using \modelGPTF{} as a base model.

In summary, our contributions are as follows:
\begin{enumerate}
    \item We introduce a framework for decomposing events into atomic facts and tracking their \interval{}s on a timeline.
    \item Based on that framework, we develop a method, \ours{}, for detecting time-aware factual contradictions in outlines. % \ours{} also serves as a plugin structure for improving factual consistency in text planning and generation.
    \item We apply \ours{} to story outline generation, defining a task and LLM-based evaluation metrics for detecting contradictions. The results confirm our approach's empirical effectiveness.
    % \item Since our method using smaller models outperforms larger, we offer an evaluation metric and salable oversight signal based on smaller models. This provides a potential avenue for improving factual consistency in the context of large model generation during pre-training.\kevin{can you just cut this last item, it's a bit muddled and a bit repetitive with the third item while also speculating on future work that you didn't actually do}
\end{enumerate}

\section{Related Work}
\paragraph{Fact Verification.} 
Fact verification is a task widely studied in natural language processing, ranging from verifying scientific 
claims \citep{thorne-etal-2018-fever, wadden-etal-2020-fact} to validating fake news \citep{wang-2017-liar, augenstein-etal-2019-multifc}. 
Unlike verifying claims against a database of facts, existing work has also demonstrated the feasibility of performing verification within context \citep{mihaylova-etal-2019-semeval, shaar-etal-2022-role, li2023contradoc}. 
We also draw inspiration from efforts to 
decompose complex sentences into multiple atomic facts using LLMs \citep{fan-etal-2020-generating, kamoi2023wice, min2023factscore}. 
Compared with prior works, our approach specifically operates on temporal structures, maintaining time-dependent fact validity intervals 
% within the extensive corpus 
to handle the \textit{dynamic} nature of the \world{} as facts change over time.
\paragraph{State Tracking.}
% The concept of state tracking in natural language processing is grounded in the principle of temporal locality, tracing its origins back to the hidden states in the structure of Recurrent Neural Networks (RNNs) \citep{medsker2001recurrent, hochreiter1997long}. 
%\kevin{not sure if you need most of this first sentence - RNNs are not very related to our work.}
Prior work on state tracking ranges from dialogue tracking \citep{thomson2010bayesian, chao2019bertdst} and memory and entities 
networks \citep{DBLP:journals/corr/SukhbaatarSWF15, henaff2017tracking} to using neural checklists \citep{kiddon-etal-2016-globally} 
and story planning \citep{DBLP:journals/corr/abs-2004-14967}. With the advent of LLMs, 
state tracking in long-form story planning and generation has shifted towards explicit tracking using natural language, 
ranging from unstructured text \citep{zhou2023recurrentgpt} to structured dictionaries \citep{yang2022re3}. Additionally, there also 
exists some work on generating better temporal fact validity by predicting the validity interval \citep{zhang-choi-2023-mitigating}, 
jointly modeling text with its timestamp \citep{dhingra-etal-2022-time}, or utilizing Wikipedia timestamps \citep{jang2023temporalwiki}. 
Our approach is related to the latter, but the main difference is our method operates on the generated output from a language model, and 
focuses on post-hoc detection rather than the calibration of a language model. 
% \zhiheng{Just delete the later sentence?}
% Our methodology occupies a nuanced position between 
% unstructured text and structured memory, by maintaining an independent collection of natural language atomic facts. 
% We thus support more flexible hypotheses 
% compared with structured memory, while offering improved interpretability relative to purely text-based memory.\kevin{this paragraph is a bit long}
% By doing the structured 
% decomposition, we observe better performance on contradiction detection tasks on story outlines compared to pure text-
% based methods like Chain of Thought \citep{wei2023chainofthought}. (See section 4) 

\paragraph{Hierarchical Generation.}
Hierarchical generation is widely applicable across various domains of long-form content creation, such as story generation. It can be implemented through the internal hidden states of the model \citep{li-etal-2015-hierarchical, shen-etal-2019-towards, DBLP:journals/corr/abs-2112-07916}, or explicitly via natural language text or structured schema \citep{yao2019planandwrite, DBLP:journals/corr/abs-2004-14967, tian-peng-2022-zero,mirowski2022cowriting, yang2022re3, yang2022doc}. The paradigm of hierarchical generation brings both benefits and challenges to our work. On one hand, it facilitates the resolution of contradictions at higher levels compared to those detected at the bottom level or during sequential generation. On the other hand, it may require a more complex data structure to maintain the validity intervals of facts.
% the non-chronological generation process (we use breadth first extension, though depth first extension also work in our method), 
%\kevin{at some point in your methods sections, whereever you say you are doing BFS, you should mention that you also trivially support DFS - emphasize that we're very flexible.} \zhiheng{Does that work?}
\section{Time-Aware Atomic Facts}
\label{sec:atomic_facts}
% Here we introduce \textbf{AtomicTrack}, a method that decomposes narratives from top to bottom into trackable atomic facts on the timeline, and utilize that to detects factual contradict in story outline. 
% \subsection{Method Description}

Existing work focuses on leveraging the semantic decomposition capabilities of LLMs for fine-grained fact-checking \citep{min2023factscore}. In this session, we propose an event-centric, time-aware atomic fact decomposition method. We track facts on a timeline, building toward the design of our \ours{} system. 
By analyzing the structure of an event and the representation of the world, we develop a method to better decompose events into atomic facts (Section 3.1) with forward and backward directions. We then discuss how to use knowledge of contradictions between atomic facts to determine the \interval{}s of these atomic facts on the timeline (Section 3.2). Through this decomposition of facts and by maintaining \interval{}s, our method is particularly effective for time-varying facts as is common in domains such as story writing. Examples illustrating the concepts in this section are provided in Appendix \ref{sec:concept_define}.
\begin{figure*}[ht]
    \centering
    \includegraphics[width=0.9\textwidth]{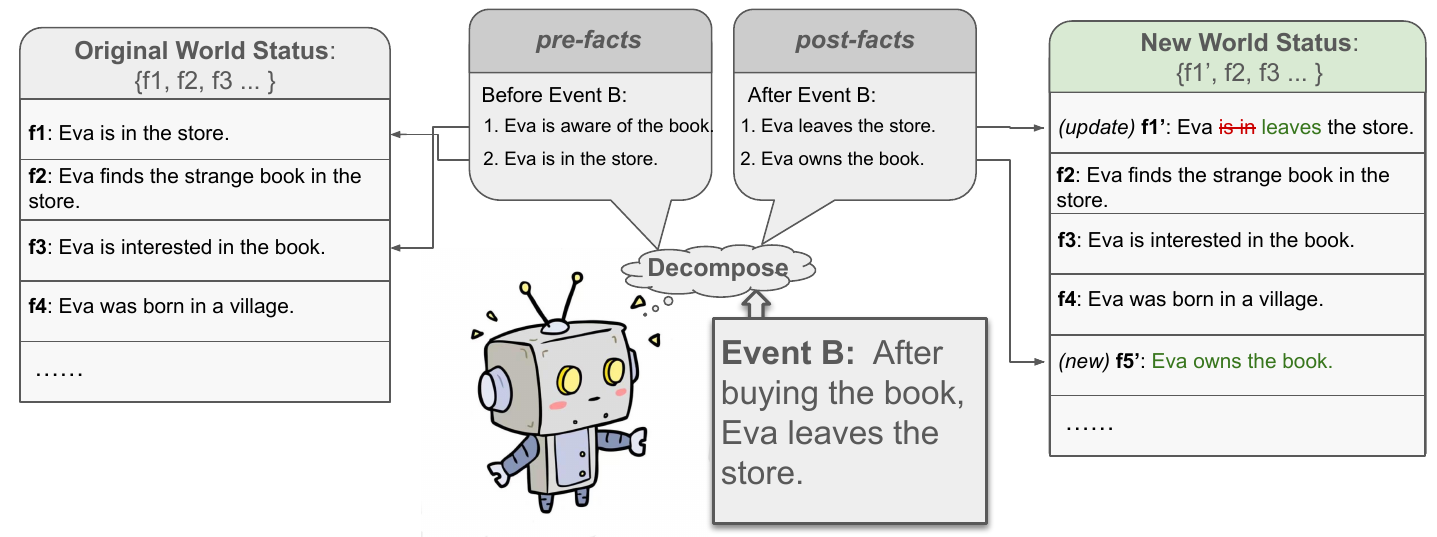}
    \caption{Decomposition of an event, to \textit{verify} and \textit{update} %\kevin{keep wording consistent - pick one of "update" or "verify" and use it throughout the paper} 
    the \world{} for the event. 
    Moving forward on the arrow of time in this example, we first retrieve all facts corresponding to \prefacts{} from 
    the \world{} to check for any conflicting fact pairs (\textit{Verification}). We then replace any 
    fact in the \world{} that contradicts a \postfact{} with the corresponding new \postfact{} (\textit{Update}).% \kevin{in this figure, end sentences with periods. "Eva knows the book" -> "Eva is aware of the book" or "Eva sees the book." also, "After Buying the Book Eva leaves the Store" -> uncapitalize all the random nouns. "Origin World Status" -> "Original World Status". "Eva is born in a village" -> "Eva was born in a village"}
    }
    \label{fig:decompose}
    \vspace{-2em}
\end{figure*}
\subsection{Fact Decomposition}
\label{sec:narration_analyze}
\paragraph{Events and Directional Atomic Facts.} 
In narrative theory, \textit{events} represent transitions in the \textit{world state}, reflecting shifts that impact characters, settings, or the overarching scenario \citep{huhn2009living}. We model these transitions using \textit{directional atomic facts}, capturing the essence of events as transformations from one state to another. Atomic fact means a basic and concise statement that conveys a single piece of information \citep{min2023factscore}, see examples in Figure \ref{fig:decompose}. Then we employ a novel strategy using LLMs to decompose these events into distinct atomic facts with directions and a time \interval{} as shown in Figure \ref{fig:decompose}. These facts are then compared pairwise to assess their interrelations and impacts on the narrative structure. Recognizing that events represent transformations in the \world{}, we classify these atomic facts into three categories: \textit{\prefacts{}}, \textit{\postfacts{}}, and \textit{\statfacts{}}. \prefacts{} are those truths that exist before an event takes place, \postfacts{} are truths that emerge following an event, and \statfacts{} are those truths that remain unchanged throughout.\footnote{Traditional AI methods like STRIPS \citep{fikes1971strips} discuss \textit{preconditions} and \textit{postconditions}. Similar to our \textit{pre-facts} and \textit{post-facts}, they model the requirements of the world state before an action and the changes to the world state after an action. We prefer the terms {pre-fact} and {post-fact} to better express our meaning of tracking time-varying natural language facts with validity intervals on a timeline.} To simplify the problem, we interpret \statfacts{} as a straightforward amalgamation of a \prefact{} and a \postfact{}, and henceforth concern ourselves with only the latter two categories. Figure \ref{fig:decompose} shows an example of how we decompose event $\mathcal{B}$ into \prefacts{} and \postfacts{}. 

% \kevin{the concept of pre-, post-, and \statfacts{} isn't trivial; it might help to have some examples. In the decomposition figure, maybe instead of before the event A, after the event A just say \prefact{} and \postfact{} so it's clear, then refer to the figure here.} 
\label{sec:fact_decomposition}

\paragraph{World State.} The term \textit{\world{}} corresponds to a set of facts that hold at a particular point in time.  %\citep{wittgenstein2023tractatus}. \zhiheng{The term `World Status' aligns with Ludwig Wittgenstein's view in the Tractatus Logico-Philosophicus that the world comprises all facts at a specific time. \citep{wittgenstein2023tractatus} Wittgenstein argues that these facts, rather than objects, form the world's structure. }
Compared with object-based tracking, fact-based tracking provides a more flexible state space. The \world{} at any given moment represents the maximum set of non-contradicting facts. For example, in a process that moves forward in time, at each new event, the \world{} is used to cross-check with \prefacts{} and is updated with \postfacts{}. If a fact in the \world{} contradicts with a new \postfact{}, then we drop the former (Figure \ref{fig:decompose}).
\begin{figure*}[ht]
    \centering
%\vspace{-1em}
\includegraphics[width=0.8\textwidth]{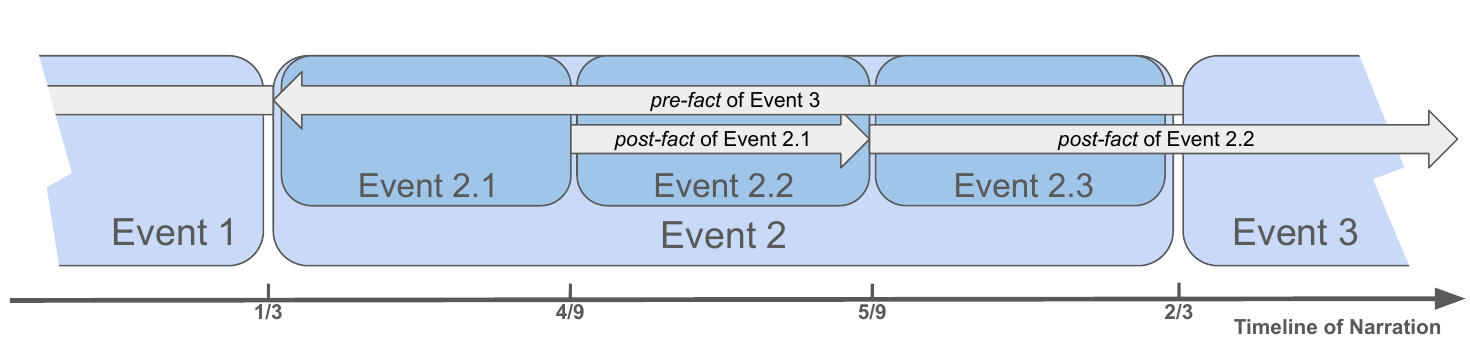}
    \caption{The timeline of narration. The start time and end time of any given event can be split recursively into sub-events. \prefacts{} begin at the left boundary and point to the left; \postfacts{} begin at the right boundary and point to the right.
    }
    \vspace{-1em}
    \label{fig:state}
\end{figure*}
\subsection{State Tracking on Timeline}
\label{sec:pseudo_timeline}
\paragraph{Event Time Interval.}
Any given event can be decomposed into multiple sub-events according to the author's desired level of detail. 
For instance, previous works use this strategy to generate a story outline by doing this decomposition recursively \citep{yang2022doc, wang-etal-2023-improving-pacing}.  
To effectively model this hierarchical temporal structure, we define the time interval of an event as a continuous segment along the narration timeline (See Figure \ref{fig:state}). 
For simplicity, we set the time interval of the entire narrative (e.g., story outline) to be $[0,1]$. When one event ($\mathcal{B}$) is a subevent of another event ($\mathcal{A}$), $\mathcal{B}$'s time interval is contained within $\mathcal{A}$'s time interval, with adjacent events possessing non-intersecting intervals. That is, for each level of a partial hierarchical outline, if we want to generate $k$ subevents, we have:
\begin{align}
\forall i &\in 1..k: \nonumber \\
l_i &= l + (i-1)\cdot\frac{r-l}{k} \tag{1} \\
r_i &= l + i\cdot\frac{r-l}{k} \tag{2}
\end{align}

\paragraph{Fact Validity Interval.}

We now consider an event with a \interval{} $[l, r]$. We assume a \prefact{} $f$ is from some time point $x$ in the event, 
i.e., $x\in[l,r]$. Then we define the \textit{\interval{}} 
of $f$ as $(-\inf, x]$. This implies $f$ is valid 
before the time point $x$, but not afterwards. Similar logic applies to \postfacts{}, where the default \interval{} is $[x, \inf)$. However, as it is unclear how to pinpoint the exact value of $x$, we simplify the problem by assigning 
the default \interval{} to be $(-\inf, l]$ for \prefacts{} and $[r, \inf)$ for \postfacts{}.
\paragraph{Update Condition.}
To handle alterations in the \world{}, we introduce a mechanism that updates the interval when two facts with the 
same direction (either both \prefacts{} or both \postfacts{}) contradict each other. The premise here is that the more ``up-to-date''
fact is deemed more reliable and reflective of the updated \world{} on the interval where they overlap. For example, consider two \postfacts{}: ``Eva is in the store'' %\kevin{remember, change these quotes to be `` and '' throughout} 
and ``Eva left the store.'' Assuming they have the \interval{}s $[\frac{4}{9}, \inf)$ and $[\frac{5}{9}, \inf)$ respectively, 
and are contradictory, we would need to adjust the first \interval{} to be $[\frac{4}{9}, \frac{5}{9})$. As exemplified in the
\postfact{} of Event 2.1 and the \postfact{} of Event 2.2 in Figure \ref{fig:state}, we will update the former fact to make the two \interval{}s not 
overlap with each other.
\begin{figure}[ht]
    \centering
    \vspace{-1.5em}
    %\hspace*{0.1cm} % 向左移动图片，可以根据需要调整数值
    \includegraphics[width=0.8\columnwidth]{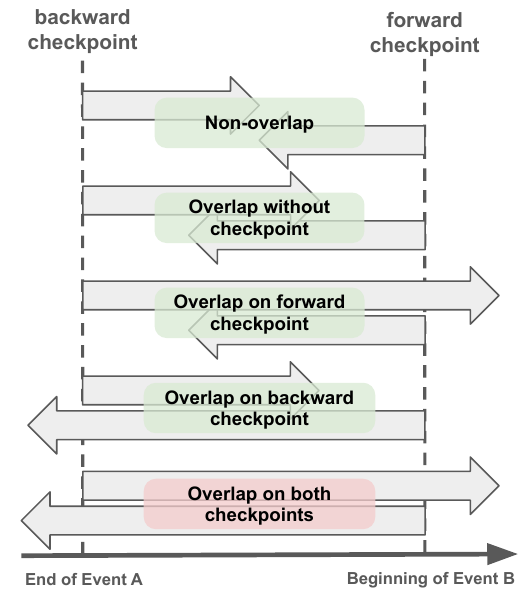} % 调整图片宽度以适应一栏
    \caption{Five possible situations for a \prefact{} and \postfact{} contradicting each other on different points or intervals,
    depending on their respective \interval{}s.
    In our implementation, we only flag a contradiction when a contradiction is detected on both checkpoints (the last situation)
    to maximize confidence in our predictions. %\kevin{Non-Overlap -> Non-overlap. Overlap without Checkpoint -> Overlap without checkpoints. Overlap on Both checkpoint -> Overlap on both checkpoints. Begin of Plot B -> Beginning of Plot B. (basically, fixing grammar and making capitalization consistent.) also, i wonder if you should use red = contradiction and green = non-contradiction, i.e. swap the colors so the first 4 boxes are green and the last is red, since in the western world often red = bad and green = good}
    % In our implementation, we only accept the facts that contradict on both checkpoints 
    % (the last situation) to ensure robustness, since the update step will potentially missed.
    }
    \vspace{-1.5em}
    \label{fig:overlap}
\end{figure}

\begin{figure*}[th]
    % \vspace{-3em}
    \centering
    \includegraphics[width=0.7\textwidth]{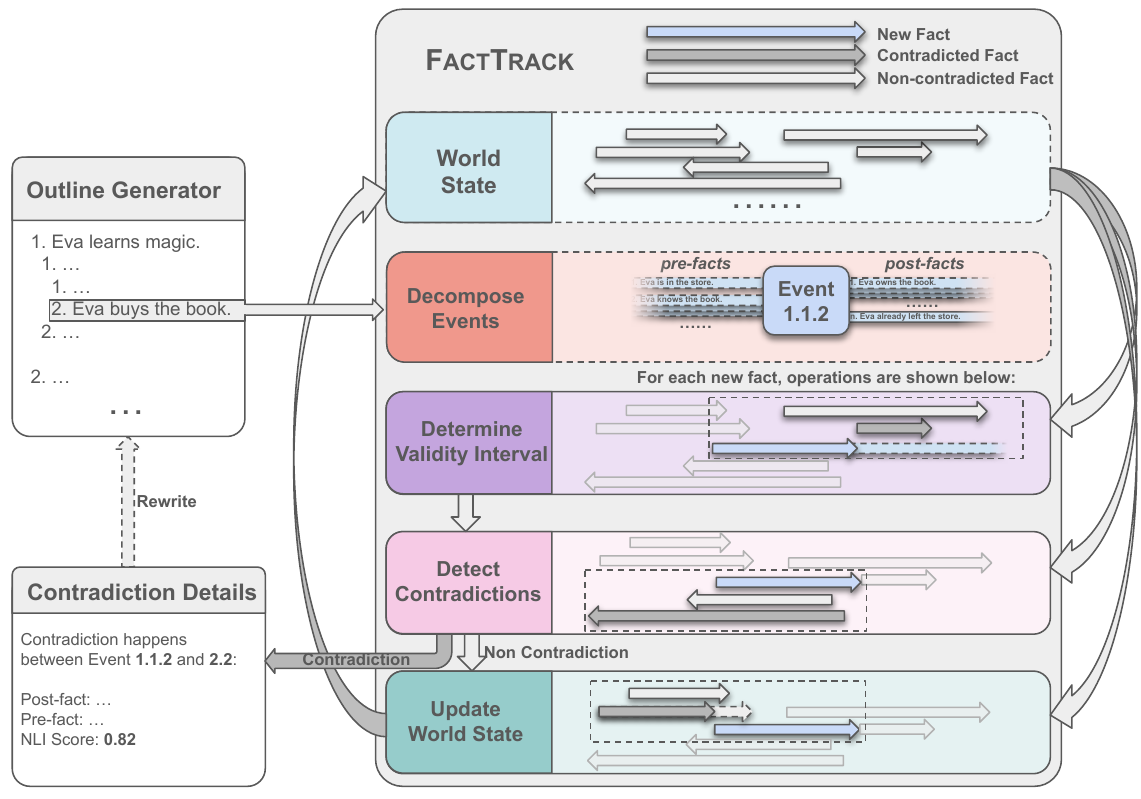}
    \caption{The general pipeline for how we maintain our data structure. 
    We begin with a new event (e.g., plot point in a story outline), which we decompose into several \prefacts{} and \postfacts{} (\textit{Decompose Events}).
    For each fact, we determine its \interval{} based on the \world{} (\textit{Determine Validity Interval}), and then detect any contradictions with existing facts in the world state (\textit{Detect Contradictions}). 
    If the fact does not contradict any existing fact in the \world{}, then we update the \world{} with the new fact (\textit{Update World State}). Otherwise, 
    we write down details about the contradiction, and rewrite the new event conditioned on the preexisting event and details about the contradiction. 
    Note that \textit{Determine Validity Interval} and \textit{Update World State} are only between facts in the same direction, while \textit{Detect Contradictions} are only between facts in different directions. 
    % \kevin{generally this figure is good, just a couple of suggestions for modifying the text: "Pick one fact" -> "World state update for one new fact shown below:", also maybe "plot 1.1.2 and 2.2" -> "Event 1.1.2 and 2.2"?}%\kevin{rewrite the outline, or just the new event? if the latter, just be specific and say rewrite the new event}
     %  \zhiheng{Should we change the wording in the pipeline? e.g. Grounding -> Determine; Verify -> Detect; Status -> State/World}\zhiheng{Write the whole thing in the box; eva buys the book, non-contradicted fact; contradiction and non contradiction; contradiction details...}
    }
    % \zhiheng{Make four stage have different color.}
    % \vspace{-1em}
    \label{fig:pipeline}
\end{figure*}

\paragraph{Contradiction Condition.}
\label{sec:contradict_condition}
We flag a contradiction between a \postfact{} from an earlier event and a \prefact{} from a later event if they 
overlap in time and contradict each other. Suppose the \interval{} of the \prefact{} is 
$(l_1, r_1]$ and the interval of \postfact{} is $[l_2, r_2)$, where $l_2<r_1$ since the \prefact{} is from the later event. 
Figure \ref{fig:overlap} shows five possible relationships between these two facts with respect to the overlap in their \interval{}s. 
 As shown in Figure \ref{fig:overlap}, a \textit{checkpoint} indicates the boundary of an event where one of the two facts begins. Although different choices are possible, in our approach, we only flag contradictions for the case 
``Overlap on both checkpoints.''%\kevin{note that when you have a quote at the end of the sentence, you're supposed to put the period inside the quote, not outside. e.g., "overlap on both checkpoints." not "overlap on both checkpoints". i fixed this case for you; you can see if there are others to fix} We discuss this choice further in Section \ref{sec:detect_contradict}. % \kevin{note you haven't defined checkpoints at this stage.}\zhiheng{But we shown on the Figure}\zhiheng{Define the checkpoint here.} \
Thus the constraint we use for flagging contradictions is: 
\begin{align}
l_1 \le l_2 \le r_1 \le r_2 \tag{3}
\end{align}
Similar ideas also can be found in Allen's Interval Algebra \citep{allen1983maintaining}, Our constraint corresponds to Allen's Interval Algebra is: $Fact_1\ o\ Fact_2$, but there is slightly different since because different operations dictate the directionality of facts as we shown in Figure \ref{fig:pipeline}.

%\ zhiheng{Move the later part to Section 4}
% In our approach, we use the strictest constraint as we shown in Figure \ref{fig:overlap}, since we only care about whether there are contradicdict on the \textit{checkpoint}, which is the $r_1$ and $l_2$, where two fact begins. Our method can also be seen as a persistent extension about the chronological fact checking. In the chronological fact checking process in Figure \ref{fig:decompose}, we always do the fact check at the beginning time point of event B, we call it \textit{forward checkpoint} since it works on the checking procedure following the arrow of time.  Symmetry, we call the $l_2$ as the \textit{backward checkpoint}. In fact, the considerations on both \textit{checkpoints} in Figure \ref{fig:overlap} provide our method more robustness for missed update step due to insufficient fact decomposition and bad performance of the retriever or NLI model. 
%\zhiheng{TODO:Refer more about figure 4}

\section{\ours{}}
%\zhiheng{Refer figure more than once, change the example in the figure}
\label{sec:maintain_data_structure}
We are now ready to introduce our method, \ours{} (Figure \ref{fig:pipeline}). 
\ours{} can be understood as a data structure for representing a \world{} (Section \ref{sec:data_structure}) coupled with a pipeline of operations for updating this data structure given a new event that we want to insert (Section \ref{sec:operation_pipeline}). Notably, events do not need to be inserted in chronological order, which is useful for hierarchical inputs such as story outlines. In our pipeline, we start by breaking down the event into several \prefacts{} and \postfacts{}. Then for each fact, we conduct a series of interval operations to determine its \interval{}, identify any contradictions, and finally update the \interval{} of facts in \world{}. We also discuss how our method recognizes contradictions between two atomic facts (Section \ref{sec:contradict_rec}).

\subsection{World State Maintaining}
\label{sec:data_structure}
As shown in the light blue block in Figure \ref{fig:pipeline}, we maintain two lists to keep track of all \prefacts{} and \postfacts{}. For each atomic fact, we store its content $s_i$, start time $t_{i, begin}$, and end time $t_{i, end}$.
%\kevin{i know you also store Contriever but i removed that from here as you haven't even introduced anything related to Contriever yet, so it'll be confusing. Contriever can be considered an implementation detail.}
% , and its Embedding using Contriever $e_{contriver, i}$\kevin{cite?} in the list. 
% The reason we use lists instead of more advanced data structures is that this step is not a time bottleneck; when access is needed, we iterate the entire list.\kevin{i think this detail for why you use list isn't very important, honestly you could just cut it}
% \zhiheng{Add the Details about how we maintain the interval}
% \label{sec:decompose}
\subsection{Operation Pipeline}
\label{sec:operation_pipeline}
As shown in Figure \ref{fig:pipeline}, our operation pipeline consists of four steps which we execute in order: 
\begin{enumerate}
    \item \textit{Decompose Events.} decompose a new event into \prefacts{} and \postfacts{}.
    \item \textit{Determine Validity Interval.} Use the \world{} to find the \interval{} for each atomic \prefact{} or \postfact{}.
    \item \textit{Detect Contradictions.} Check if the current fact's \interval{} contradict with existing facts.
    \item \textit{Update World State.} If needed, update existing facts as necessary and add the current fact to the \world{}.
\end{enumerate}
Pseudocode is shown in Appendix \ref{app:pseudo_code}.

\subsubsection{Decompose Events} In the orange block in Figure \ref{fig:pipeline}, we decompose the event into \prefacts{}, \postfacts{}, and static facts. We use zero-shot prompting with an LLM and parse the output for structure afterward; see details in Appendix \ref{app:prompt_decomposition}. After this step, we execute the following three steps in sequence for each atomic fact.

\subsubsection{Determine Validity Interval}
Given a new atomic fact, we use all the facts in the \world{} to determine its \interval{} (purple block in Figure \ref{fig:pipeline}). Taking a post-fact as an example, our initial default \interval{} is $[l, \inf)$, where $l$ is the right boundary of the event it came from. We then check the facts in the \world{} whose left boundary is greater than $l$, in order from left to right, until we find the first contradiction. Then, we set the right boundary of the current fact as the left boundary of the detected fact and return.

\subsubsection{Detect Contradictions}
As shown in the pink block in Figure \ref{fig:pipeline}, we retrieve all facts that overlap with the current fact, and check for contradictions. Note that this process can operate forward in time, as shown in Figure \ref{fig:decompose}, as well as backward along the arrow of time. There are multiple ways to define the ``overlap'' of two directional segments. In our approach, we use the strictest constraint in Figure \ref{fig:overlap}, since we only care about whether there are contradictions on the \textit{checkpoints}, where the two directional facts begin. By requiring contradiction on both checkpoints, as depicted in Figure \ref{fig:overlap}, our method gains better robustness against errors in fact decomposition or the retrieval process, eliminating many false positives. After this step, if there is a contradiction detected, we can print the contradiction information as shown in the examples in Appendix \ref{app:example}, otherwise we directly update the \world{}. 

\label{sec:detect_contradict}
\subsubsection{Update World State} 
As shown in the green block in Figure \ref{fig:pipeline}, if we decide to accept the event when no contradiction is detected, we will use this step to update the \world{}. We first use the current fact to update the \interval{}s of existing facts in the \world{}. Taking a post-fact with interval $[l, r)$ as an example, to update the \interval{}s, we retrieve all facts $[L, R)$ satisfying $L\le l \le R \le r$. If those two facts contradict, then we update the \interval{} $[L, R)$ to be $[L, l)$. Finally, we add the new fact $[l, r)$ to the \world{}.

\subsection{Contradiction Recognition}
\label{sec:contradict_rec}
We use an NLI model finetuned on outputs annotated by GPT4 (Appendix \ref{app:nli}). By using that model, we can give every pair of facts a contradiction score from 0 to 1 corresponding to how likely they are to contradict each other. If this score is over a set threshold, we say that the two facts contradict in our method. For the update condition and contradiction condition, we use different thresholds since false negatives for the update condition are less harmful than for the contradiction condition---it may not cause any major problems if we miss an update to a fact's \interval{}, but we do not want to ignore an actual contradiction. Additionally, we also use a retrieval model as a filter to reduce the computational cost, see filtering details in Appendix \ref{app:pseudo_code}. 

%\kevin{maybe point to an appendix where you discuss the details of this filtering}

\begin{table*}[t]
\centering \small
% \vspace{-3em}
\begin{tabular}{lcccc}
\toprule
Method & \metricPairwise{}  & \metricContext{}\\
\midrule
\text{\baselineOutline{} (\modelGPTF{})} & 2.355 \(\pm\) 0.163 & {2.859} \(\pm\) 0.149 \\
\text{\ours{} (\modelLLaMA{}, top 300)} & {2.393} \(\pm\) 0.164 & 2.777 \(\pm\) 0.146 \\
\text{\ours{} (\modelGPTF{}, top 300}) & \textbf{2.599} \(\pm\) 0.148 & \textbf{3.133} \(\pm\) 0.123 \\
\hline
\text{Random} & 1.419 \(\pm\) 0.075 & \text{1.62 \(\pm\) 0.087} \\
\text{\baselinePairwise{} (\modelLLaMA{})} & 1.452 \(\pm\) 0.080 & 1.64 \(\pm\) 0.088 \\
\text{\baselineOutline{} (\modelGPTT{})} & 1.556 \(\pm\) 0.068 & 1.902 \(\pm\) 0.078 \\
\text{\ours{} (\modelLLaMA{}, random 500)} & \textbf{1.836} \(\pm\) 0.092 & \textbf{2.046} \(\pm\) 0.092 \\
\bottomrule
\end{tabular}
\caption{Comparison results of \ours{} (based on \modelLLaMA{}) against corresponding baselines on 90 depth-3 story outlines. The contradiction scores range from 1 to 5, as annotated by \modelGPTF{}; higher is better. \metricPairwise{} indicates that the \modelGPTF{} evaluator only sees the events corresponding to the two facts being checked, while in \metricContext{}, \modelGPTF{} sees the whole story outline as context. We downsample \ours{}'s detections either randomly (e.g., random 500) or based on top contradiction scores to match the number of detections with baselines for fairer comparison; see Appendices \ref{app:prompt_metrics} and \ref{app:exp_stat} for experiment details. \ours{} based on \modelLLaMA{} achieves performance comparable to GPT4-Turbo while significantly outperforming other baselines. Additionally, when \ours{} is run on GPT4, the performance surpasses all baselines.}
%\vspace{-1.5em}
\label{tab:annotation}
\end{table*}

\begin{table*}[t]
\centering \small
%\vspace{-3em}
\begin{tabular}{lcccc}
\toprule
Method & \metricPairwise{}  & \metricContext{}\\
\midrule
\text{CoT (\textit{without decompose events})} & 2.355 & 2.859\\
\text{\ours{}} & \textbf{2.599} & \textbf{3.133}  \\
\hline
\text{\ours{} (\textit{without track facts}}) & 2.380 & 2.364\\
\text{\ours{}} & \textbf{2.412} & \textbf{2.386}  \\
\hline
% \text{\ours{} (NLI Model, maximum 300)} & {2.393} & {2.777}  \\
\text{\ours{} (few-shot GPT4)} & \textbf{2.212} & \textbf{2.216}  \\
\text{\ours{} (few-shot LLaMA-7B-Chat)} & {2.186} & {2.196}  \\
\text{\ours{} (NLI Model, default)} & {1.836} & {2.046}  \\
\bottomrule
\end{tabular}
\caption{Ablations on individual components of \ours{}, including removing \textit{Decompose Events}, \textit{Track Facts} (combining \textit{Determine Validity Interval} and \textit{Update World State}), and replacing \textit{Detect Contradictions} with other methods. Both \textit{Decompose Events} and \textit{Track Facts} are critical to performance, while \textit{Detect Contradictions} can be improved by changing its internal modules and thresholds.}
% \vspace{-2em}
\label{tab:ablation}
\end{table*}

\section{Evaluation}
% \zhiheng{Add more evaluation results when our method can do, but GPT4 not}
\paragraph{Experiment Setup.} To examine our method's empirical effectiveness, we apply \ours{} to the task of detecting the contradictions in a story outline. While our method can also work online, detecting problems during the outline generation process and providing feedback, we evaluate only offline in this work. We employ an outline generator similar to the detailed outliner from \citet{yang2022doc}, modifying the prompt to include more factual information (Appendix \ref{app:prompt_baseline}). In particular, we follow their paradigm of breadth-first outline expansion. 
To balance length and knowledge density, we use story outlines with three layers, with each event having three sub-events, for 39 events in total. 
We estimate there are 3-5 true contradictions per outline on average (Appendix \ref{app:outline_gen}).
In the evaluation process, we randomly select 90 premises from the WritingPrompts dataset \citep{fan-etal-2018-hierarchical} and use \modelLLaMA{} \citep{touvron2023llama2} to generate the outlines; note that we run \ours{} on \modelLLaMA{} as well.
The task can be understood as a detection task: given the outline, we want to retrieve some candidate event pairs, indicating that the model considers them to be contradictory. Outline statistics are shown in Appendix \ref{app:experiment_stat}.

% \vspace{-2em}
{
\small
\begin{align*}
\text{Input}:&\  [premise, event_1, event_{1.1}, \cdots event_{3.3.3}]\\
\text{Output}:&\ [(id_{1, 1}. id_{1, 2}), \cdots, (id_{n, 1}, id_{n, 2})]
\end{align*}
}
% \vspace{-3em}
\paragraph{Baselines.} With no directly applicable prior methods, we designed two baselines:
\begin{enumerate}
    \item \textbf{\baselinePairwise{} on \modelLLaMA{}}: The model compares two retrieved events to make predictions of contradictions. This approach scales well but struggles with maintaining temporal validity.
    \item \textbf{\baselineOutline{} on {GPT} series}: The model reviews the entire text using Chain of Thought (CoT) to identify contradictory event nodes. Limited by context window size, this method processes an outline of 39 events under our setting, averaging about 4800 tokens, using \modelGPTT{} and \modelGPTF{}.
\end{enumerate}
Due to varying numbers of positive detections, results were downsampled (see Appendix \ref{app:exp_stat}).
% \vspace{-1em}

\paragraph{Metrics.} Due to the complexity and context-dependent nature of annotating event contradictions, we haven't found a clear method to label ground truth data, even with human input (see Appendix \ref{app:human_annotation}). Upon review, \modelGPTF{} \citep{openai2023gpt4} annotations proved to be higher quality and less noisy. Since contradictions aren't binary and can be nuanced and depend on context, we score them from 1 to 5 (Appendix \ref{app:prompt_metrics}). We evaluate methods by the average contradiction score in detected pairs—a higher score indicates better detection. We developed two metric variations using \modelGPTF{} context:

\vspace{-0.5em}
\begin{enumerate}
    \item \textbf{\metricPairwise{}} directly labels contradictions by examining pairs of events directly.
    \item \textbf{\metricContext{}} labels event pairs within the full outline context.
\end{enumerate}
\vspace{-0.5em}
Classical metrics aren't applicable without gold labels, but we estimate precision and recall using our metrics in Appendix \ref{app:exp_stat}.

\paragraph{Evaluation Results.} Table \ref{tab:annotation} shows the results of our experiment. \ours{} on \modelLLaMA{} is significantly better on the two metrics than both \baselineOutline{} using \modelGPTT{} and \baselinePairwise{} on \modelLLaMA{}. We also achieve comparable performance with GPT4-Turbo, despite only using \modelLLaMA{} in our method. When we run \ours{} on GPT4, the performance significantly surpasses all baselines\footnote{Note that while we used \modelGPTF{} to generate the fact contradiction data to finetune our NLI model to work on more complex text, \modelGPTF{} is not directly used in the pipeline of \ours{} on on \modelLLaMA{}. See details in Appendix \ref{app:nli}.}
These results confirm that \ours{} effectively enhances contradiction detection in story planning by decomposing events and maintaining validity intervals of atomic facts. Examples are in Appendix \ref{app:example}.

Qualitative inspection shows that event decomposition remains a bottleneck, as language ambiguity can lead to misunderstandings before decomposition. Appendix \ref{app:badcase} illustrates a failure case due to ambiguity. Additionally, our method only detects binary contradictions and doesn't address more complex scenarios. Detailed error analysis is in Appendix \ref{app:badcase}.

\subsection{Ablation Study}
We examine the roles and alternatives of the modules in \ours{}: \textit{Decompose Events}, \textit{Track Facts} (\textit{Determine Validity Interval} $+$ \textit{Update World State}), and \textit{Detect Contradictions}.

\begin{enumerate}
\item \textit{Without Decompose Events}: To assess performance without the \textit{Decompose Events} module, we substitute it with the Chain of Thought, as detailed in Table \ref{tab:prompt_baseline2}.
\item \textit{Without Track Facts}: This variant excludes the \textit{Determine Validity Interval} and \textit{Update World State} steps, treating all facts as valid by default.
\item \textit{Replacing the NLI Model in Detecting Contradictions}: We use the NLI model to lower the pipeline's computational cost. We explore replacing it with few-shot LLMs and estimate the potential improvements quantitatively.
\end{enumerate} 

\paragraph{Results}
Table \ref{tab:ablation} shows that both the \textit{Decompose Events} and \textit{Track Facts} modules are critical to performance. However, the \textit{Detect Contradictions} module can be improved by using few-shot versions of GPT-4 and LLaMA-7B-Chat, suggesting that more fine-grained annotation and knowledge distillation could enhance performance further. This indicates that while \ours{} has already substantially outperformed the baselines, there is potential for even better performance.

\subsection{Document-level contradict detection}
We extended the application of \ours{} to the ContraDoc dataset \cite{li2023contradoc}, a dataset for evaluating document-level contradictions. In alignment with the original study, we employed a binary judgment framework, which involves directly prompting large language models (LLMs) as a baseline. Our approach, \ours{}, demonstrated a better F1-score compared to the baseline models. The observed reduction in precision relative to the baseline can be attributed to the presence of borderline contradictions within the documents and the ambiguity in the decomposition process. The results are detailed in Table \ref{tab:contra_doc}.

\begin{table}[t]
\centering \small 
\begin{tabularx}{\columnwidth}{lXXXX}
\toprule
\textbf{Experiment} & \textbf{Precision} & \textbf{Recall} & \textbf{F1} \\
\midrule
GPT4o-mini & \textbf{89.29\%} & 5.57\% & 10.48\% \\
\ours{} & 52.71\% & \textbf{62.81\%} & \textbf{57.32\%}\\
\bottomrule
\end{tabularx}
\caption{Performance of the baseline and \ours{} in the Binary Judgement experiment in ContraDoc. \ours{} was implemented using GPT4o-mini, with splitting events by sentences conducted via the nltk package.}
\vspace{-1.5em}
\label{tab:contra_doc}
\end{table}
\vspace{0.5em}

\section{Conclusion and Future Work}
As the complexity of texts generated by LLMs increases, understanding their structures and tracking time-varying factual information becomes an increasingly important bottleneck in long-form generation.
In this work, we have introduced \ours{}, a fact-tracking framework that decomposes events into pre-facts and post-facts and maps them onto a timeline, facilitating the tracking of narrative progress and detecting contradictions between events.
Experimental results show that when we apply \ours{} to contradiction detection on structured story outlines using \modelLLaMA{} as the base model, our method achieves performance comparable to GPT4 and significantly outperforms baselines using the same base LLM. Furthermore, when we run our method on GPT4, we find that its performance significantly outperforms all baselines.

By effectively maintaining factual consistency over extended contexts, we envision \ours{} serving not only as a fact-tracking module for complex content planning but also as an efficient automatic evaluation metric for contradictions in extensive texts.
We additionally envision several further possibilities for improving our system, such as enhancing decomposition accuracy, structuring atomic facts more effectively (for instance, based on entities), and/or maintaining timelines in narratives that do not follow a strict chronological order.
Moreover, although we only experiment on story outlines in this work, \ours{} is in principle generally applicable to other domains, and we hope that our framework's capacity for managing time-specific knowledge could be of use in other areas as well, such as detecting fake news and dynamically updating knowledge bases.

% \kevin{remember that speculation about future applications you had at the end of your contributions in the intro? here's where it should go instead. you can talk about how you hope that \ours{} can inform and contribute to future story planning methods to both evaluate and help improve them with respect to maintaining factual consistency over long context.}

% \newpage
\section*{Limitations}
The difficulty of identifying all contradictions and partial contradictions significantly constrained our evaluations (see Appendix \ref{app:human_anno}), limiting us to obtaining gold labels of contradictions. Therefore, we used GPT-4 to annotate the contradicting score as a proxy for the evaluation task.

The context window size of baseline and evaluation metrics also restricted us from running experiments on outlines much longer than the current 2000-3000 words. While \ours{} is capable of detecting contradictions in such outlines with near-linear cost growth, we did not evaluate them in this work due to the potential performance degradation of GPT-4 in longer contexts.

Additionally, the challenge of conducting thorough evaluations impacted the system's development. Many decisions, such as prompt design, the choice of models for each module (e.g., the few-shot language model outperforms the NLI model as shown in Table \ref{tab:ablation}), and the selection of hyperparameters (e.g., thresholds for the contriver and NLI models), were made manually rather than through rigorous validation. As a result, there may be considerable room for improvement in the detailed design of individual modules.

Finally, \ours{}'s performance may decrease on LLMs that lack strong generation and instruction-following capabilities.

% 1. 依赖大语言模型的生成能力，对模型的生成能力和instruction following能力变化敏感，e.g.用gpt4替代llama2-7B-chat后观察到了大幅度性能的提升
% 2. 由于gold label的获取困难，没有直接测量事实跟踪的精确度，而是选择矛盾检测作为替代任务；以及就是没有进行人类标注
% 3. The difficulty of careful evaluation also affected system development. Many system design choices (e.g., prompt design, model used in each module) and hyperparameters (e.g., threshold of contriver and NLI model) are simply selected manually, rather than chosen based on careful validation. Thus it is likely that substantial room for improvement remains in the detailed design of our individual modules. 
\section*{Ethics Statement}\label{sec:ethic}
Since \ours{} is built upon existing LLMs, we may inherit any potential biases and harms from those systems. However, in \ours{}, we focus on tracking facts and detecting factual inconsistencies during the process of creative story generation, with an emphasis on the interpretability of the world state within narrative structures. Our focus on factuality limits the potential abuse of language models and may be a useful tool for mitigating such abuses in the first place.

\ours{} is also currently designed only for English, although translating our prompts to other languages shouldn't be difficult in principle. However, performance might suffer in lower-resource languages, depending on the base LLM.

\section*{Reproducibility Statement} \label{sec:reproduct} 
We saved all intermediate computation results, including our NLI dataset as well as results from the decomposition and inference steps. The finetuning process of the NLI model is also described in the appendix; otherwise, we use existing open-source or API-based LLMs for inference with temperature 0. Thus we believe our work is highly reproducible, and all of our code and data will be open-sourced upon publication. 
However, as we use the OpenAI API in some of our experimental comparisons, we cannot rule out the possibility of minor differences in inference results due to future API updates. 

\bibliography{fact_track}

\begin{thebibliography}{41}
\expandafter\ifx\csname natexlab\endcsname\relax\def\natexlab#1{#1}\fi

\bibitem[{Allen(1983)}]{allen1983maintaining}
James~F Allen. 1983.
\newblock Maintaining knowledge about temporal intervals.
\newblock \emph{Communications of the ACM}, 26(11):832--843.

\bibitem[{Augenstein et~al.(2019)Augenstein, Lioma, Wang, Chaves~Lima, Hansen,
  Hansen, and Simonsen}]{augenstein-etal-2019-multifc}
Isabelle Augenstein, Christina Lioma, Dongsheng Wang, Lucas Chaves~Lima, Casper
  Hansen, Christian Hansen, and Jakob~Grue Simonsen. 2019.
\newblock \href {https://doi.org/10.18653/v1/D19-1475} {{M}ulti{FC}: A
  real-world multi-domain dataset for evidence-based fact checking of claims}.
\newblock In \emph{Proceedings of the 2019 Conference on Empirical Methods in
  Natural Language Processing and the 9th International Joint Conference on
  Natural Language Processing (EMNLP-IJCNLP)}, pages 4685--4697, Hong Kong,
  China. Association for Computational Linguistics.

\bibitem[{Cai et~al.(2021)Cai, Zhang, Huang, Lam, and Dolan}]{cai2021narrative}
Deng Cai, Yizhe Zhang, Yichen Huang, Wai Lam, and Bill Dolan. 2021.
\newblock \href {http://arxiv.org/abs/2012.11157} {Narrative incoherence
  detection}.

\bibitem[{Chao and Lane(2019)}]{chao2019bertdst}
Guan-Lin Chao and Ian Lane. 2019.
\newblock \href {http://arxiv.org/abs/1907.03040} {Bert-dst: Scalable
  end-to-end dialogue state tracking with bidirectional encoder representations
  from transformer}.

\bibitem[{de~Marneffe et~al.(2008)de~Marneffe, Rafferty, and
  Manning}]{de-marneffe-etal-2008-finding}
Marie-Catherine de~Marneffe, Anna~N. Rafferty, and Christopher~D. Manning.
  2008.
\newblock \href {https://aclanthology.org/P08-1118} {Finding contradictions in
  text}.
\newblock In \emph{Proceedings of ACL-08: HLT}, pages 1039--1047, Columbus,
  Ohio. Association for Computational Linguistics.

\bibitem[{Dhingra et~al.(2022)Dhingra, Cole, Eisenschlos, Gillick, Eisenstein,
  and Cohen}]{dhingra-etal-2022-time}
Bhuwan Dhingra, Jeremy~R. Cole, Julian~Martin Eisenschlos, Daniel Gillick,
  Jacob Eisenstein, and William~W. Cohen. 2022.
\newblock \href {https://doi.org/10.1162/tacl_a_00459} {Time-aware language
  models as temporal knowledge bases}.
\newblock \emph{Transactions of the Association for Computational Linguistics},
  10:257--273.

\bibitem[{Fan et~al.(2018)Fan, Lewis, and Dauphin}]{fan-etal-2018-hierarchical}
Angela Fan, Mike Lewis, and Yann Dauphin. 2018.
\newblock \href {https://doi.org/10.18653/v1/P18-1082} {Hierarchical neural
  story generation}.
\newblock In \emph{Proceedings of the 56th Annual Meeting of the Association
  for Computational Linguistics (Volume 1: Long Papers)}, pages 889--898,
  Melbourne, Australia. Association for Computational Linguistics.

\bibitem[{Fan et~al.(2020)Fan, Piktus, Petroni, Wenzek, Saeidi, Vlachos,
  Bordes, and Riedel}]{fan-etal-2020-generating}
Angela Fan, Aleksandra Piktus, Fabio Petroni, Guillaume Wenzek, Marzieh Saeidi,
  Andreas Vlachos, Antoine Bordes, and Sebastian Riedel. 2020.
\newblock \href {https://doi.org/10.18653/v1/2020.emnlp-main.580} {Generating
  fact checking briefs}.
\newblock In \emph{Proceedings of the 2020 Conference on Empirical Methods in
  Natural Language Processing (EMNLP)}, pages 7147--7161, Online. Association
  for Computational Linguistics.

\bibitem[{Fikes and Nilsson(1971)}]{fikes1971strips}
Richard~E Fikes and Nils~J Nilsson. 1971.
\newblock Strips: A new approach to the application of theorem proving to
  problem solving.
\newblock \emph{Artificial intelligence}, 2(3-4):189--208.

\bibitem[{Guo et~al.(2021)Guo, Ainslie, Uthus, Onta{\~{n}}{\'{o}}n, Ni, Sung,
  and Yang}]{DBLP:journals/corr/abs-2112-07916}
Mandy Guo, Joshua Ainslie, David~C. Uthus, Santiago Onta{\~{n}}{\'{o}}n, Jianmo
  Ni, Yun{-}Hsuan Sung, and Yinfei Yang. 2021.
\newblock \href {http://arxiv.org/abs/2112.07916} {Longt5: Efficient
  text-to-text transformer for long sequences}.
\newblock \emph{CoRR}, abs/2112.07916.

\bibitem[{Henaff et~al.(2017)Henaff, Weston, Szlam, Bordes, and
  LeCun}]{henaff2017tracking}
Mikael Henaff, Jason Weston, Arthur Szlam, Antoine Bordes, and Yann LeCun.
  2017.
\newblock \href {http://arxiv.org/abs/1612.03969} {Tracking the world state
  with recurrent entity networks}.

\bibitem[{H{\"u}hn et~al.(2009)H{\"u}hn, Meister, Pier, Schmid, and
  Sch{\"o}nert}]{huhn2009living}
Peter H{\"u}hn, Jan~Christoph Meister, John Pier, Wolf Schmid, and J{\"o}rg
  Sch{\"o}nert. 2009.
\newblock \href {https://www-archiv.fdm.uni-hamburg.de/lhn/node/39.html} {The
  living handbook of narratology}.
\newblock \emph{Hamburg: Hamburg University. URL: http://www.lhn.uni-hamburg.de
  (Retrieved on 12.03. 2020)}.

\bibitem[{Izacard et~al.(2021)Izacard, Caron, Hosseini, Riedel, Bojanowski,
  Joulin, and Grave}]{izacard2021contriever}
Gautier Izacard, Mathilde Caron, Lucas Hosseini, Sebastian Riedel, Piotr
  Bojanowski, Armand Joulin, and Edouard Grave. 2021.
\newblock \href {https://doi.org/10.48550/ARXIV.2112.09118} {Unsupervised dense
  information retrieval with contrastive learning}.

\bibitem[{Jang et~al.(2023)Jang, Ye, Lee, Yang, Shin, Han, Kim, and
  Seo}]{jang2023temporalwiki}
Joel Jang, Seonghyeon Ye, Changho Lee, Sohee Yang, Joongbo Shin, Janghoon Han,
  Gyeonghun Kim, and Minjoon Seo. 2023.
\newblock \href {http://arxiv.org/abs/2204.14211} {Temporalwiki: A lifelong
  benchmark for training and evaluating ever-evolving language models}.

\bibitem[{Kamoi et~al.(2023)Kamoi, Goyal, Rodriguez, and
  Durrett}]{kamoi2023wice}
Ryo Kamoi, Tanya Goyal, Juan~Diego Rodriguez, and Greg Durrett. 2023.
\newblock \href {http://arxiv.org/abs/2303.01432} {Wice: Real-world entailment
  for claims in wikipedia}.

\bibitem[{Kiddon et~al.(2016)Kiddon, Zettlemoyer, and
  Choi}]{kiddon-etal-2016-globally}
Chlo{\'e} Kiddon, Luke Zettlemoyer, and Yejin Choi. 2016.
\newblock \href {https://doi.org/10.18653/v1/D16-1032} {Globally coherent text
  generation with neural checklist models}.
\newblock In \emph{Proceedings of the 2016 Conference on Empirical Methods in
  Natural Language Processing}, pages 329--339, Austin, Texas. Association for
  Computational Linguistics.

\bibitem[{Li et~al.(2023)Li, Raheja, and Kumar}]{li2023contradoc}
Jierui Li, Vipul Raheja, and Dhruv Kumar. 2023.
\newblock Contradoc: Understanding self-contradictions in documents with large
  language models.
\newblock \emph{arXiv preprint arXiv:2311.09182}.

\bibitem[{Li et~al.(2015)Li, Luong, and Jurafsky}]{li-etal-2015-hierarchical}
Jiwei Li, Thang Luong, and Dan Jurafsky. 2015.
\newblock \href {https://doi.org/10.3115/v1/P15-1107} {A hierarchical neural
  autoencoder for paragraphs and documents}.
\newblock In \emph{Proceedings of the 53rd Annual Meeting of the Association
  for Computational Linguistics and the 7th International Joint Conference on
  Natural Language Processing (Volume 1: Long Papers)}, pages 1106--1115,
  Beijing, China. Association for Computational Linguistics.

\bibitem[{Liu et~al.(2023)Liu, Lin, Hewitt, Paranjape, Bevilacqua, Petroni, and
  Liang}]{liu2023lost}
Nelson~F. Liu, Kevin Lin, John Hewitt, Ashwin Paranjape, Michele Bevilacqua,
  Fabio Petroni, and Percy Liang. 2023.
\newblock \href {http://arxiv.org/abs/2307.03172} {Lost in the middle: How
  language models use long contexts}.

\bibitem[{Mihaylova et~al.(2019)Mihaylova, Karadzhov, Atanasova, Baly,
  Mohtarami, and Nakov}]{mihaylova-etal-2019-semeval}
Tsvetomila Mihaylova, Georgi Karadzhov, Pepa Atanasova, Ramy Baly, Mitra
  Mohtarami, and Preslav Nakov. 2019.
\newblock \href {https://doi.org/10.18653/v1/S19-2149} {{S}em{E}val-2019 task
  8: Fact checking in community question answering forums}.
\newblock In \emph{Proceedings of the 13th International Workshop on Semantic
  Evaluation}, pages 860--869, Minneapolis, Minnesota, USA. Association for
  Computational Linguistics.

\bibitem[{Min et~al.(2023)Min, Krishna, Lyu, Lewis, Yih, Koh, Iyyer,
  Zettlemoyer, and Hajishirzi}]{min2023factscore}
Sewon Min, Kalpesh Krishna, Xinxi Lyu, Mike Lewis, Wen-tau Yih, Pang~Wei Koh,
  Mohit Iyyer, Luke Zettlemoyer, and Hannaneh Hajishirzi. 2023.
\newblock Factscore: Fine-grained atomic evaluation of factual precision in
  long form text generation.
\newblock \emph{arXiv preprint arXiv:2305.14251}.

\bibitem[{Mirowski et~al.(2022)Mirowski, Mathewson, Pittman, and
  Evans}]{mirowski2022cowriting}
Piotr Mirowski, Kory~W. Mathewson, Jaylen Pittman, and Richard Evans. 2022.
\newblock \href {http://arxiv.org/abs/2209.14958} {Co-writing screenplays and
  theatre scripts with language models: An evaluation by industry
  professionals}.

\bibitem[{OpenAI(2023)}]{openai2023gpt4}
OpenAI. 2023.
\newblock \href {http://arxiv.org/abs/2303.08774} {Gpt-4 technical report}.

\bibitem[{Ouyang et~al.(2022)Ouyang, Wu, Jiang, Almeida, Wainwright, Mishkin,
  Zhang, Agarwal, Slama, Ray, Schulman, Hilton, Kelton, Miller, Simens, Askell,
  Welinder, Christiano, Leike, and Lowe}]{ouyang2022training}
Long Ouyang, Jeff Wu, Xu~Jiang, Diogo Almeida, Carroll~L. Wainwright, Pamela
  Mishkin, Chong Zhang, Sandhini Agarwal, Katarina Slama, Alex Ray, John
  Schulman, Jacob Hilton, Fraser Kelton, Luke Miller, Maddie Simens, Amanda
  Askell, Peter Welinder, Paul Christiano, Jan Leike, and Ryan Lowe. 2022.
\newblock \href {http://arxiv.org/abs/2203.02155} {Training language models to
  follow instructions with human feedback}.

\bibitem[{Pei et~al.(2022)Pei, Ananthasubramaniam, Wang, Zhou, Dedeloudis,
  Sargent, and Jurgens}]{pei-etal-2022-potato}
Jiaxin Pei, Aparna Ananthasubramaniam, Xingyao Wang, Naitian Zhou, Apostolos
  Dedeloudis, Jackson Sargent, and David Jurgens. 2022.
\newblock \href {https://doi.org/10.18653/v1/2022.emnlp-demos.33} {{POTATO}:
  The portable text annotation tool}.
\newblock In \emph{Proceedings of the 2022 Conference on Empirical Methods in
  Natural Language Processing: System Demonstrations}, pages 327--337, Abu
  Dhabi, UAE. Association for Computational Linguistics.

\bibitem[{Rashkin et~al.(2020)Rashkin, Celikyilmaz, Choi, and
  Gao}]{DBLP:journals/corr/abs-2004-14967}
Hannah Rashkin, Asli Celikyilmaz, Yejin Choi, and Jianfeng Gao. 2020.
\newblock \href {http://arxiv.org/abs/2004.14967} {Plotmachines:
  Outline-conditioned generation with dynamic plot state tracking}.
\newblock \emph{CoRR}, abs/2004.14967.

\bibitem[{Shaar et~al.(2022)Shaar, Alam, Da~San~Martino, and
  Nakov}]{shaar-etal-2022-role}
Shaden Shaar, Firoj Alam, Giovanni Da~San~Martino, and Preslav Nakov. 2022.
\newblock \href {https://doi.org/10.18653/v1/2022.findings-naacl.122} {The role
  of context in detecting previously fact-checked claims}.
\newblock In \emph{Findings of the Association for Computational Linguistics:
  NAACL 2022}, pages 1619--1631, Seattle, United States. Association for
  Computational Linguistics.

\bibitem[{Shen et~al.(2019)Shen, Celikyilmaz, Zhang, Chen, Wang, Gao, and
  Carin}]{shen-etal-2019-towards}
Dinghan Shen, Asli Celikyilmaz, Yizhe Zhang, Liqun Chen, Xin Wang, Jianfeng
  Gao, and Lawrence Carin. 2019.
\newblock \href {https://doi.org/10.18653/v1/P19-1200} {Towards generating long
  and coherent text with multi-level latent variable models}.
\newblock In \emph{Proceedings of the 57th Annual Meeting of the Association
  for Computational Linguistics}, pages 2079--2089, Florence, Italy.
  Association for Computational Linguistics.

\bibitem[{Sukhbaatar et~al.(2015)Sukhbaatar, Szlam, Weston, and
  Fergus}]{DBLP:journals/corr/SukhbaatarSWF15}
Sainbayar Sukhbaatar, Arthur Szlam, Jason Weston, and Rob Fergus. 2015.
\newblock \href {http://arxiv.org/abs/1503.08895} {Weakly supervised memory
  networks}.
\newblock \emph{CoRR}, abs/1503.08895.

\bibitem[{Thomson and Young(2010)}]{thomson2010bayesian}
Blaise Thomson and Steve Young. 2010.
\newblock Bayesian update of dialogue state: A pomdp framework for spoken
  dialogue systems.
\newblock \emph{Computer Speech \& Language}, 24(4):562--588.

\bibitem[{Thorne et~al.(2018)Thorne, Vlachos, Christodoulopoulos, and
  Mittal}]{thorne-etal-2018-fever}
James Thorne, Andreas Vlachos, Christos Christodoulopoulos, and Arpit Mittal.
  2018.
\newblock \href {https://doi.org/10.18653/v1/N18-1074} {{FEVER}: a large-scale
  dataset for fact extraction and {VER}ification}.
\newblock In \emph{Proceedings of the 2018 Conference of the North {A}merican
  Chapter of the Association for Computational Linguistics: Human Language
  Technologies, Volume 1 (Long Papers)}, pages 809--819, New Orleans,
  Louisiana. Association for Computational Linguistics.

\bibitem[{Tian and Peng(2022)}]{tian-peng-2022-zero}
Yufei Tian and Nanyun Peng. 2022.
\newblock \href {https://doi.org/10.18653/v1/2022.naacl-main.262} {Zero-shot
  sonnet generation with discourse-level planning and aesthetics features}.
\newblock In \emph{Proceedings of the 2022 Conference of the North American
  Chapter of the Association for Computational Linguistics: Human Language
  Technologies}, pages 3587--3597, Seattle, United States. Association for
  Computational Linguistics.

\bibitem[{Touvron et~al.(2023)Touvron, Martin, Stone, Albert, Almahairi,
  Babaei, Bashlykov, Batra, Bhargava, Bhosale, Bikel, Blecher, Ferrer, Chen,
  Cucurull, Esiobu, Fernandes, Fu, Fu, Fuller, Gao, Goswami, Goyal, Hartshorn,
  Hosseini, Hou, Inan, Kardas, Kerkez, Khabsa, Kloumann, Korenev, Koura,
  Lachaux, Lavril, Lee, Liskovich, Lu, Mao, Martinet, Mihaylov, Mishra,
  Molybog, Nie, Poulton, Reizenstein, Rungta, Saladi, Schelten, Silva, Smith,
  Subramanian, Tan, Tang, Taylor, Williams, Kuan, Xu, Yan, Zarov, Zhang, Fan,
  Kambadur, Narang, Rodriguez, Stojnic, Edunov, and
  Scialom}]{touvron2023llama2}
Hugo Touvron, Louis Martin, Kevin Stone, Peter Albert, Amjad Almahairi, Yasmine
  Babaei, Nikolay Bashlykov, Soumya Batra, Prajjwal Bhargava, Shruti Bhosale,
  Dan Bikel, Lukas Blecher, Cristian~Canton Ferrer, Moya Chen, Guillem
  Cucurull, David Esiobu, Jude Fernandes, Jeremy Fu, Wenyin Fu, Brian Fuller,
  Cynthia Gao, Vedanuj Goswami, Naman Goyal, Anthony Hartshorn, Saghar
  Hosseini, Rui Hou, Hakan Inan, Marcin Kardas, Viktor Kerkez, Madian Khabsa,
  Isabel Kloumann, Artem Korenev, Punit~Singh Koura, Marie-Anne Lachaux,
  Thibaut Lavril, Jenya Lee, Diana Liskovich, Yinghai Lu, Yuning Mao, Xavier
  Martinet, Todor Mihaylov, Pushkar Mishra, Igor Molybog, Yixin Nie, Andrew
  Poulton, Jeremy Reizenstein, Rashi Rungta, Kalyan Saladi, Alan Schelten, Ruan
  Silva, Eric~Michael Smith, Ranjan Subramanian, Xiaoqing~Ellen Tan, Binh Tang,
  Ross Taylor, Adina Williams, Jian~Xiang Kuan, Puxin Xu, Zheng Yan, Iliyan
  Zarov, Yuchen Zhang, Angela Fan, Melanie Kambadur, Sharan Narang, Aurelien
  Rodriguez, Robert Stojnic, Sergey Edunov, and Thomas Scialom. 2023.
\newblock \href {http://arxiv.org/abs/2307.09288} {Llama 2: Open foundation and
  fine-tuned chat models}.

\bibitem[{Wadden et~al.(2020)Wadden, Lin, Lo, Wang, van Zuylen, Cohan, and
  Hajishirzi}]{wadden-etal-2020-fact}
David Wadden, Shanchuan Lin, Kyle Lo, Lucy~Lu Wang, Madeleine van Zuylen, Arman
  Cohan, and Hannaneh Hajishirzi. 2020.
\newblock \href {https://doi.org/10.18653/v1/2020.emnlp-main.609} {Fact or
  fiction: Verifying scientific claims}.
\newblock In \emph{Proceedings of the 2020 Conference on Empirical Methods in
  Natural Language Processing (EMNLP)}, pages 7534--7550, Online. Association
  for Computational Linguistics.

\bibitem[{Wang(2017)}]{wang-2017-liar}
William~Yang Wang. 2017.
\newblock \href {https://doi.org/10.18653/v1/P17-2067} {{``}liar, liar pants on
  fire{''}: A new benchmark dataset for fake news detection}.
\newblock In \emph{Proceedings of the 55th Annual Meeting of the Association
  for Computational Linguistics (Volume 2: Short Papers)}, pages 422--426,
  Vancouver, Canada. Association for Computational Linguistics.

\bibitem[{Wang et~al.(2023)Wang, Yang, Liu, and
  Klein}]{wang-etal-2023-improving-pacing}
Yichen Wang, Kevin Yang, Xiaoming Liu, and Dan Klein. 2023.
\newblock \href {https://doi.org/10.18653/v1/2023.findings-emnlp.723}
  {Improving pacing in long-form story planning}.
\newblock In \emph{Findings of the Association for Computational Linguistics:
  EMNLP 2023}, pages 10788--10845, Singapore. Association for Computational
  Linguistics.

\bibitem[{Yang et~al.(2022{\natexlab{a}})Yang, Klein, Peng, and
  Tian}]{yang2022doc}
Kevin Yang, Dan Klein, Nanyun Peng, and Yuandong Tian. 2022{\natexlab{a}}.
\newblock Doc: Improving long story coherence with detailed outline control.
\newblock \emph{arXiv preprint arXiv:2212.10077}.

\bibitem[{Yang et~al.(2022{\natexlab{b}})Yang, Peng, Tian, and
  Klein}]{yang2022re3}
Kevin Yang, Nanyun Peng, Yuandong Tian, and Dan Klein. 2022{\natexlab{b}}.
\newblock Re3: Generating longer stories with recursive reprompting and
  revision.
\newblock \emph{arXiv preprint arXiv:2210.06774}.

\bibitem[{Yao et~al.(2019)Yao, Peng, Weischedel, Knight, Zhao, and
  Yan}]{yao2019planandwrite}
Lili Yao, Nanyun Peng, Ralph Weischedel, Kevin Knight, Dongyan Zhao, and Rui
  Yan. 2019.
\newblock \href {http://arxiv.org/abs/1811.05701} {Plan-and-write: Towards
  better automatic storytelling}.

\bibitem[{Zhang and Choi(2023)}]{zhang-choi-2023-mitigating}
Michael Zhang and Eunsol Choi. 2023.
\newblock \href {https://doi.org/10.18653/v1/2023.emnlp-main.879} {Mitigating
  temporal misalignment by discarding outdated facts}.
\newblock In \emph{Proceedings of the 2023 Conference on Empirical Methods in
  Natural Language Processing}, pages 14213--14226, Singapore. Association for
  Computational Linguistics.

\bibitem[{Zhou et~al.(2023)Zhou, Jiang, Cui, Wang, Xiao, Hou, Cotterell, and
  Sachan}]{zhou2023recurrentgpt}
Wangchunshu Zhou, Yuchen~Eleanor Jiang, Peng Cui, Tiannan Wang, Zhenxin Xiao,
  Yifan Hou, Ryan Cotterell, and Mrinmaya Sachan. 2023.
\newblock \href {http://arxiv.org/abs/2305.13304} {Recurrentgpt: Interactive
  generation of (arbitrarily) long text}.

\end{thebibliography}

\appendix

\newpage
\section{Concept Clarification}
\label{sec:concept_define}
\subsection{Time-aware Contradiction}
A \textbf{time-aware contradiction} occurs when two factual statements remain contradictory even when temporal context is considered. For instance, the statements:
\begin{itemize}
    \item \textit{``Donald Trump is the president of the United States.''}
    \item \textit{``Joe Biden is the president of the United States.''}
\end{itemize}
may appear contradictory. However, when interpreted with temporal context—Trump serving before 2021 and Biden serving from 2021 onward—the contradiction is resolved. Thus, this is \emph{not} a time-aware contradiction. Instead, a time-aware contradiction persists even after considering temporal information.

\subsection{Time-dependent Fact}
A \textbf{time-dependent fact} is a fact that holds true within a specific time interval but may become invalid afterward. For example:
\begin{quote}
    \textit{``Joe Biden is the president of the United States from 2021 to 2024.''}
\end{quote}
is a time-dependent fact since its validity is constrained to a specific period. In our FactTrack setting, we maintain a \emph{relative timeline} based on event order rather than absolute timestamps, allowing for contextual fact verification without requiring explicit temporal annotations.

\subsection{Dynamic Nature of the World State}
The \textbf{dynamic nature of the world state} refers to the updates in the state of entities following an event. Consider the following scenario:
\begin{quote}
    \textit{``After buying the book, Eva leaves the store.''}
\end{quote}
This event results in state changes:
\begin{itemize}
    \item \textbf{Eva's location:} ``Eva is in the store'' $\rightarrow$ ``Eva is not in the store.''
    \item \textbf{Book ownership:} ``Eva does not own the book'' $\rightarrow$ ``Eva owns the book.''
\end{itemize}
Capturing such transitions is essential for factuality assessment, as ignoring these updates can lead to incorrect conclusions about entity states.

\section{Methodology Details}
\subsection{Prompt of event Decomposition}
See Table \ref{tab:prompt_decomposition} for more detail.
\label{app:prompt_decomposition}
\begin{table*}[ht]
    \centering
    \begin{tabular}{|p{\textwidth}|}
    \hline
    Deconstruct the given event point into atomic facts, considering facts valid until before the event event (pre-facts), facts valid starting after the event event (post-facts), and facts that remain valid throughout the event (static facts). For pre-facts, identify the conditions that are present before the event, but change as a result of it. For post-facts, identify the conditions that are valid after the event, which are essentially the transformed versions of the corresponding pre-facts. Static facts are the conditions that remain true throughout the event. Please be sure to present facts as assertive statements, rather than speculative or suggestive ones.  \\
    \\
    event point: \textbf{\{event\_point\_text\}} \\
    \\
    Pre-Facts: \\
    \text{[pre-facts]}\\\\
    Post-Facts: \\
    \text{[post-facts]}\\\\
    Static Facts:\\
    \text{[static facts]} \\\\
    \hline
    \end{tabular}
    \caption{The prompt we use to decompose the event into different directional atomic facts.\label{tab:prompt_decomposition}}
\end{table*}
\subsection{Epsilon Padding on Timeline}
Note that to easily distinguish the boundary between different events, we set a tiny number $\epsilon = 10^{-6}$ as the padding between the events. Therefore, the real format of the event time interval is:
\begin{align*}
\epsilon &= 10^{-6}\\
\forall i&\in 1..k:\\
l_i &= l + (i-1)\cdot\frac{r-l-(k+1)\cdot\epsilon}{k} + \epsilon \cdot i\\
r_i &= l + i\cdot\frac{r-l}{k} + \epsilon \cdot i
\end{align*}
Figure \ref{fig:state} shows there is a slight gap between different events, and fact begins from the corresponding boundary of the event.
\subsection{Pseudocode of Interval Operation}
In our data structure
\footnote{
\ifarxiv
Refer to this file for the core part of our data structure: \url{https://github.com/cogito233/fact-track/blob/main/fact-track/core/contradict_detector.py}.
\else
Refer to for the core part of our data structure in the files submited to the submission system: \url{fact-track/core/contradict_detector.py}
\fi
}, We use a list to store the intervals of all fact content, their validity intervals, and their embeddings, using Contriever \citep{izacard2021contriever} as a retrieval model to filter out irrelevant fact pairs. Here are some definitions of the notation in the pseudocode:
\begin{enumerate}
    \item For $isOverlap(l, r, L, R)$ in the pseudocode, we use the condition $l\le L \le r\le R$, as discussed in Section \ref{sec:contradict_condition}.
    \item $Filter(p, P)$, used to sample relevant event pairs, is implemented as $sim_{contriver}$$(p, P)>0.5$.
    \item For $Same(p, P)$, as considered in the update condition, we drop identical atomic facts to reduce computation, as judged by $sim_{contriver}(p, P)>0.95$. 
    \item For $Contradict(p, P)$, we use different thresholds to make detection more effective and to make the update more robust. In the determine and update validity interval step, this means $NLI_Score>0.8$, while in the check contradiction step, this means $NLI_Score>0.2359$, as discussed in Appendix \ref{app:nli}.
\end{enumerate}
\label{app:pseudo_code}
\begin{algorithm}
\caption{Determine validity interval for a pre-fact}\label{alg:get}
\begin{algorithmic}
\Require $\text{\world{}: } W = [\forall i, (p_i, l_i, r_i)]$
\Require $\text{pre-fact: } P, \text{init time: }T$
\Ensure $\text{validity interval: } (L, R]$
\State $L, R \gets -\inf, T$
\State $W \gets Sort(W, r_i>r_j)$
\For{$(\text{prefact:}p, l, r)\in W, r<R$}
\If{Same(p, P)}
    \State $L \gets r$
    \State \textbf{break}
\EndIf
\If{Filter(p, P) $\And$ Contradict(p, P)}
    \State $L \gets r$
    \State \textbf{break}
\EndIf
\EndFor
\end{algorithmic}
\end{algorithm}
\begin{algorithm}
\caption{Verify whether current fact contradicts \world{}}\label{alg:check}
\begin{algorithmic}
\Require $\text{\world{}: } W = [\forall i, (p_i, l_i, r_i)]$
\Require $\text{pre-fact: } P, \text{validity interval: }(L, R]$
\Ensure $\text{whether there is a contradiction: } flag$
\State $L, R \gets -\inf, T$
\State $W' \gets [\ ]$
\State $flag \gets False$
\For{$(\text{post-fact:}p, l, r)\in W$}
\If{isOverlap(l, r, L, R) $\And$ Filter(p, P)}
\If{Contradict(p, P)}
    \State $flag = True$
\EndIf
\EndIf
\EndFor
\end{algorithmic}
\end{algorithm}
\begin{algorithm}
\caption{Update \world{} for a pre-fact}\label{alg:update}
\begin{algorithmic}
\Require $\text{\world{}: } W = [\forall i, (p_i, l_i, r_i)]$
\Require $\text{pre-fact: } P, \text{validity interval: }(L, R]$
\Ensure $\text{new \world{}: } W' = [\forall i, (p_i, l_i, r_i)]$
\State $L, R \gets -\inf, T$
\State $W' \gets Sort(W, l_i<l_j)$
    \For{$(\text{pre-fact:}p, l, r)\in W', l<R \And R<r$}
    \If{Same(p, P)}
        \State $L \gets r$
    \EndIf
    \If{Filter(p, P) $\And$ Contradict(p, P)}
        \State $l \gets R$
    \EndIf
\EndFor
\State $W' \gets W'.add(P, L, R)$
\end{algorithmic}
\end{algorithm}

\subsection{Finetuning NLI Models} 
Since the task of recognizing whether two atomic facts contradict is similar to NLI, we use an NLI model from Hugging Face (\texttt{MoritzLaurer/DeBERTa-v3-large-mn li-fever-anli-ling-wanli}) and fine-tune it using \modelGPTF{} annotations to adapt to the narrative domain. See the prompt in Table \ref{tab:prompt_classification}.
To construct the dataset, we generate 60 outlines from the WritingPrompts dataset \cite{fan-etal-2018-hierarchical} and use 50 for training and 10 for testing. In the training set, we have a total of 856,748 fact pairs. To create a subsample, we first use an NLI model without any finetuning to retrieve data. Then, we employ a second NLI model, which has been finetuned based on the data retrieved by the first, non-finetuned model. From the output of both models, we subsample 10,000 fact pairs for each model. As the vast majority of fact pairs are not contradictory, to improve class balance, we randomly select 5k from the top 1\% most confidently predicted contradictions for each model; 2k from the top 1-10\%; and 3k from the remaining. After removing duplicate data, we get 18,702 fact pairs, which are then annotated by \modelGPTF{} using the prompt shown in Table \ref{tab:prompt_classification}. 

Based on this annotation result, we estimate the overall positive rate of all 856,748 fact pairs as $2.98\%$. To evaluate, we re-rank the fact pairs in the test set, and divide the top 10\% of data into 10 intervals, with 100 data points randomly sampled in each interval to observe the positive rate. Therefore, we set the percentile threshold for flagging a contradiction in \ours{} at 3\% ($NLI\_Score \ge 0.2359$), with precision = 60\% and recall = 60\% on our \modelGPTF{}-annotated test set. \label{app:nli}
\begin{table}[ht]
    \centering
    \begin{tabular}{|p{\columnwidth}|}
    \hline
    Do the following statements contradict each other? Answer ``Yes'' or ``No''.\\\\
    \textbf{\{fact1\}}\\
    \textbf{\{fact2\}}\\
    \hline
    \end{tabular}
    \caption{The prompt given to \modelGPTF{} to classify two fact statements as contradicting each other. We use the result to fine-tune our NLI model to better fit the narrative domain.\label{tab:prompt_classification}}
\end{table}

\subsection{Outline Generation}
\label{app:outline_gen}
% 1. partial outline的定义
% 2. depth and bandwidth
% 3. Change compared with DOCAdopting a similar setting with slightly different prompts (see Table \ref{tab:prompt_generation}) to the Detailed Outliner approach \cite{yang2022doc}, our procedure initiates the expansion process from higher-level events, given that they encapsulate more general story information. This breadth-first strategy prioritizes the generation of higher-level events before their lower-level counterparts, making it easier to fix potential problems by editing or resampling higher-level elements. Our method can also be easily adapted to alternative methods such as depth-first search (DFS).

In our experiments, we use depth=3 and branching factor=3 to generate outlines, where depth refers to the maximum number of hierarchical outline layers, and branching factor refers to the number of sub-events into which one node decomposes.
We chose these settings because the length of such an outline (around 4800 tokens for \modelGPTF{}) makes it sufficiently complex while still somewhat tractable for annotation.
\ours{} can easily scale to a more complex outline.
Compared with the Detailed Outliner in the DOC paper, we have two changes. (1) We add the sections ``begin event'' and      ``end event'' to each event, because we want the Outline to be richer in facts and more precise in context; (2) We set the "partial outline" (prompt context for generating each new sub-event) to always be the content's direct parent, whereas in Detailed Outliners, the default setting is to include all ancestors and their siblings. This is mainly because we consider our method as using constant text window size for outline generation. Additionally, in our code (to be open-sourced upon publication), we have implemented all the options to conveniently change the above settings, which we believe will facilitate further evaluation of new methods in the future.
Our prompts are shown in Table \ref{tab:prompt_generation}. 

Based on rough estimation, we expect that outlines generated in this manner will contain approximately $3$ to $5$ gold contradictions per outline on average, as well as around several dozen partial contradictions.

\begin{table}[ht]
    \centering
    \begin{tabular}{|p{\columnwidth}|}
        \hline
        \textbf{\{partial\_outline\}}\\
        \\
        Can you break down point \textbf{\{idx\}} into up to \textbf{\{bandwidth\}} independent, chronological and similarly-scoped sub-points? Also list the names of characters that appear. Please follow the template below. Include detailed information about each character in the "Main event".  Do not answer anything else.\\
    \\
    point \textbf{\{idx\}}.1\\
    Main event: [event event]\\
    Characters: [character names]\\
    Begin event: [begin event]\\
    End event: [end event]\\
    \\
    point \textbf{\{idx\}}.2\\
    Main event: [event event]\\
    Characters: [character names] \\
    Begin event: [begin event]\\
    End event: [end event]\\
    \\
    
        ...\\
        \hline
    \end{tabular}
    \caption{The prompt was given to \modelGPTF{} to generate the outline. Compared to DOC, we added boundary events to make the generation more stable and increase fact density.\label{tab:prompt_generation}}
\end{table}
\section{Baselines and Evaluation Settings}
\subsection{Human Annotation Attempt}
\label{app:human_anno}
\begin{figure}[ht]
    \centering
    %\hspace*{0.1cm} % 向左移动图片，可以根据需要调整数值
    \includegraphics[width=\columnwidth]{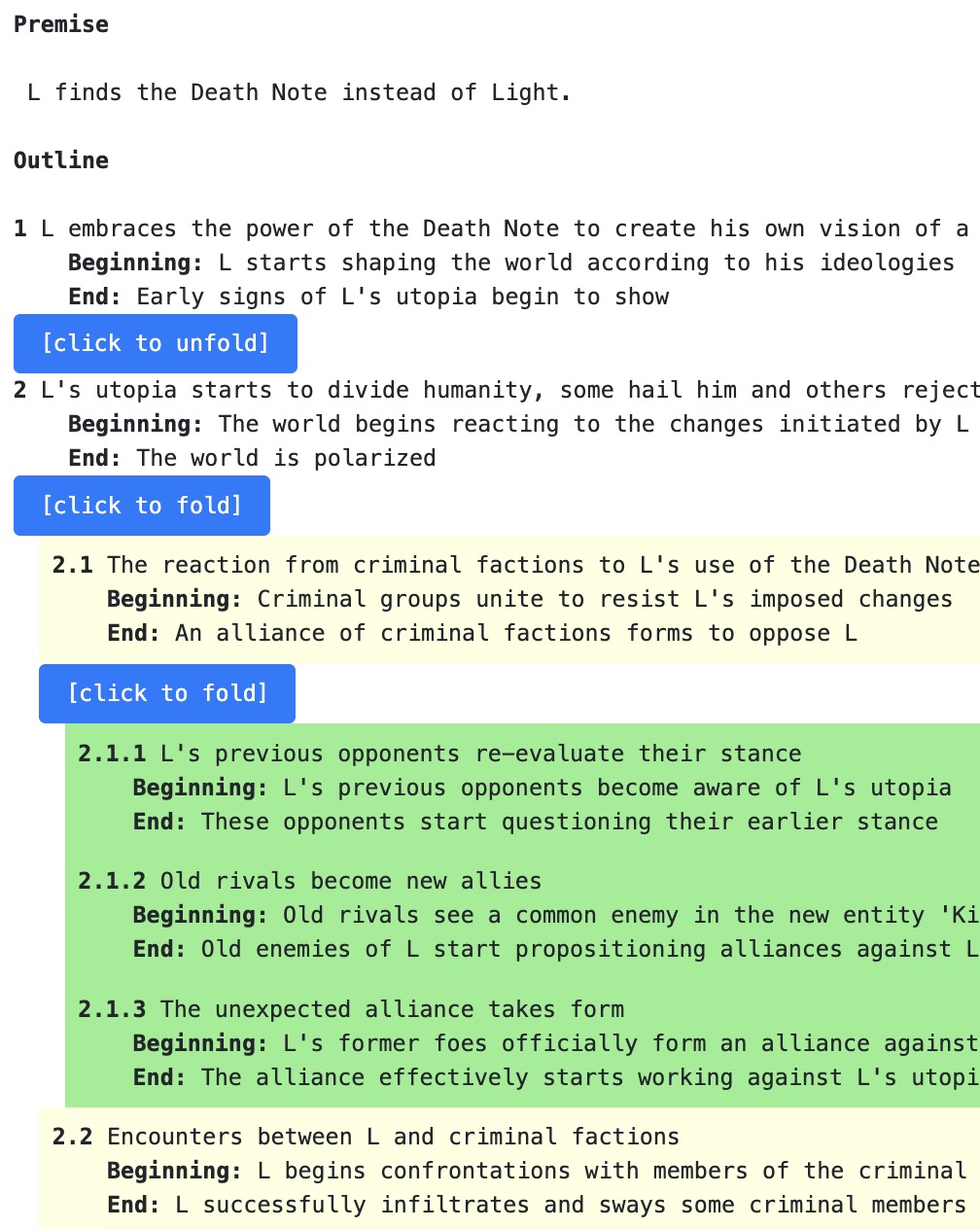} % 调整图片宽度以适应一栏
    \caption{Outline Shown in Human Annotation}
    \label{fig:humanAnno_outline}
\end{figure}

\begin{figure}[ht]
    \centering
    %\hspace*{0.1cm} % 向左移动图片，可以根据需要调整数值
    \includegraphics[width=\columnwidth]{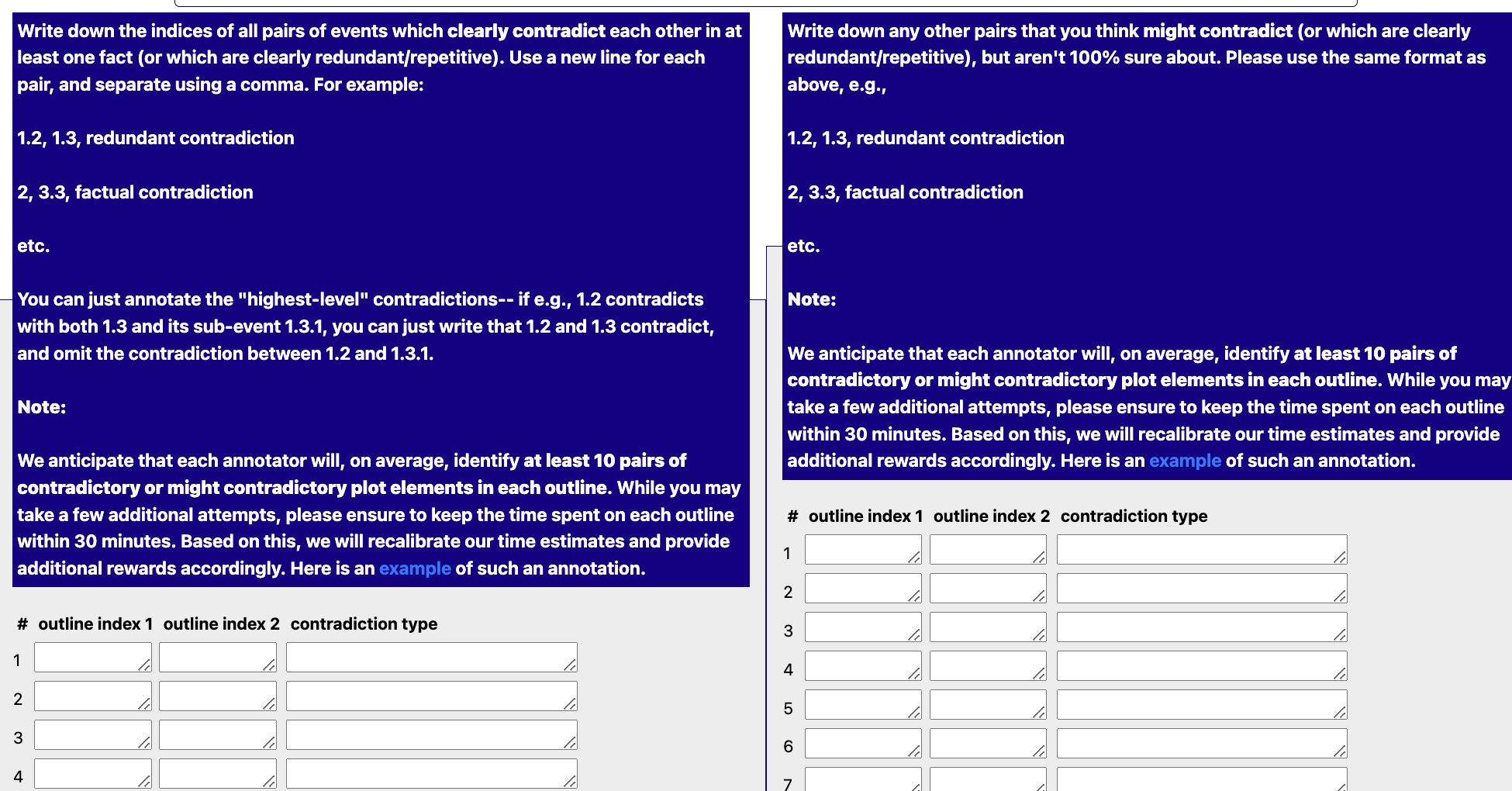} % 调整图片宽度以适应一栏
    \caption{question bank in human annotation}
    \label{fig:humanAnno_qb}
\end{figure}

We initially attempted human annotation but found the task to be highly challenging.
Even experienced human annotators and advanced LLMs struggled to accurately obtain ground truth labels.
This situation has become increasingly common in the era of LLMs. 
We used \url{prolific.co} as our human annotation platform and implemented a foldable outline with the Potato Annotation Framework \cite{pei-etal-2022-potato}, along with some structured input formats (see Figures \ref{fig:humanAnno_outline} and \ref{fig:humanAnno_qb}). The budget for each annotator was 30 minutes per Outline, at 15 USD per hour. The prompts given to human annotators were also used to establish a Full Context Detection Baseline for GPT-4, which is referred to in Table \ref{tab:prompt_baseline2}. 
In this context, we encountered problems such as: (1) annotators struggling to understand the format of the question; (2) 30 minutes being insufficient for annotators to retrieve all ground truth labels; and (3) some annotators attempting to cheat by using GPT-4 for annotation.
It is worth noting that it is theoretically feasible for human annotators to complete all the annotation tasks mentioned in this text. However, after our initial exploration, we concluded that sufficiently reducing annotation noise and achieving statistically significant conclusions would result in astonishing economic costs and probability infeasible. Therefore, in this project, we use the most advanced version of \modelGPTF{} to complete all annotation tasks.

We found that using \modelGPTF{}, with the same instructions as a human annotator, achieved better results compared to humans upon manual inspection.
However, we also found on inspection that even when our method \ours{} was deployed on \modelLLaMA{}, it maintained similar or even higher quality than \modelGPTF{}. Using a worse baseline to evaluate our method was unreasonable, so we ultimately decided to use a scoring-based metric that considers individual pairs of facts rather than full outlines. (We also tried using \modelGPTF{} for preference annotation but found it to be noisy. This was due to different events potentially contradicting each other in various aspects, making it difficult to determine which was ``better'' or ``worse.'')

\label{app:human_annotation}
\subsection{Prompts Used in Baseline}
See Table \ref{tab:prompt_baseline1} and Table \ref{tab:prompt_baseline2} for more detail.
\begin{table}[ht]
    \centering
    \begin{tabular}{|p{\columnwidth}|}
    \hline
    Question: Does those two time-ordered event point contradict each other? Answer ``Yes'' or ``No''.\\\\
    \textbf{\{event1\_text\}}\\
    \textbf{\{event2\_text\}}\\
    \hline
    \end{tabular}
    \caption{Prompt of \baselinePairwise{}, we run it on \modelLLaMA \label{tab:prompt_baseline1}}
\end{table}
\begin{table*}
    \centering \small
    \begin{tabular}{|p{\textwidth}|}
    \hline
We’re a group of AI researchers aiming to improve models' abilities to detect contradictions in story outline. We will show you a story outline below and ask you to annotate contradictions(including redundancy/repetition of the same events multiple times, or contradictory facts between two events). Please take into account that the outline is structured as a tree. In this tree-like structure, individual points such as 1.1, 1.2, and 1.3 are child nodes of event point 1, so there is no contradiction between a node with its ancestors such as 1.3 and 1. If the text you enter doesn't match our guidelines, we'll highlight the text box in bold red to alert you.\\\\

Please be as comprehensive as possible; many pairs may contradict in only one aspect but are otherwise fine. Here are some examples\\\\

Example 1\\
event 1: John is taken aback by Linda's words and prepares to respond.\\
event 2: John hears Linda and decides to answer.\\
Label: redundancy contradiction (redundancy)\\\\

Example 2\\
event 1: John's news shocks Linda, and she's unsure how to react.\\
event 2: Linda reacts with confusion and frustration.\\
Label: factual contradiction (factual: "unsure how to react" vs "reacts")\\\\

Example 3\\
event 1: John starts responding to Sarah.\\
event 2: John tells Sarah he won't comply with her demand.\\
Label: redundancy contradiction (redundancy)\\\\

Example 4\\
event 1: Ghosts lead to discovery of Max's evidence. Initially, they leave clues at crime sites. Ultimately, authorities find the same items in Max's home and arrest him.\\
event 2: Ghosts disrupt Max's routine. Initially, they alter his routine. Ultimately, his deviations expose him to the authorities.\\
Label: factual contradiction (factual: "immediate arrest" vs "exposure")\\\\
    \textbf{\{{outline}\}}\\\\

Write down the indices of all pairs of events which clearly contradict each other in at least one fact. Use a new line for each pair, and separate using a comma. For example:\\\\

factual contradiction | 1.2 | 1.3 | [Analyze: 1.2 mention that communication with Earth already established, but 1.3 indicate it is need be establish] | [Reason: the fact whether communication is established is contradictory] | [is contradiction? (Yes)]\\
factual contradiction | 2 | 3.3 | [Analyze: 2 mention that linda is angry when she grip an item, 3.3 mention that linda get into angry when shu trun around] | [Reason: based on the temperal order, it seems two diffrent events, independent and not contradict] | [is contradiction? (No)]\\
factual contradiction | 3.2 | 3.3 | [Analyze: 3.2 ends with the group locating the artifact hidden deep within the school, while 3.3 begins with the group already knowing how to deactivate the artifact, suggesting they have already located it] | [Reason: Since 3.2 happened before 3.3, so it is possible to find before deactivate it.] | [is contradiction? (No)]\\
etc.\\\\

You can just annotate the "highest-level" contradictions-- if e.g., 1.2 contradicts with both 1.3 and its sub-event 1.3.1, you can just write that 1.2 and 1.3 contradict, and omit the contradiction between 1.2 and 1.3.1.\\\\

Note:\\\\

We anticipate that each annotator will, on average, identify at least 20 pairs of contradictory or might contradictory event elements in each outline. \\
    \hline
    \end{tabular}
    \caption{Prompt of \baselineOutline{}. We run it on GPT Series models because \modelLLaMA \ can not follow the instructions reliably. \label{tab:prompt_baseline2}}
\end{table*}

\label{app:prompt_baseline}
\subsection{Prompts Used in Metrics}
\label{app:prompt_metrics}
See Table \ref{tab:prompt_metric1} and Table \ref{tab:prompt_metric2} for more detail.
\begin{table*}[ht]
    \centering
    \begin{tabular}{|p{\textwidth}|}
    \hline
    \textbf{\{event1\_text\}}\\
    \textbf{\{event2\_text\}}\\\\
    
Do these two given event in a story contain event redundant or factual inconsistency (they are assumed to be happenn on different stage in story since the index is not overlap)? Simply score the event's redundant and factual inconsistency level using one score from 1 (lowest) to 5 (highest).\\\\

Please simpliy answer:\\\\

Score of Redundancy and Factual Inconsistency: [TODO, a number from 1 to 5]\\
    \hline
    \end{tabular}
    \caption{Prompt of \metricPairwise\label{tab:prompt_metric1}}
\end{table*}
\begin{table*}[ht]
    \centering
    \begin{tabular}{|p{\textwidth}|}
    \hline
Consider the following story outline written by a ai assistant, the outline follows a tree structure, for example, the son of node 1 is 1.1, 1.2, 1.3 respectively. The outline is as follows:\\\\
\textbf{\{outline\_text\}}\\\\
Consider the following two events:\\\\
    \textbf{\{event1\_text\}}\\
    \textbf{\{event2\_text\}}\\\\
    
Do these two given event in a story contain event redundant or factual inconsistency (they are assumed to be happenn on different stage in story since the index is not overlap)? Simply score the event's redundant and factual inconsistency level using one score from 1 (lowest) to 5 (highest).\\\\

Please simpliy answer:\\\\

Score of Redundancy and Factual Inconsistency: [TODO, a number from 1 to 5]\\
    \hline
    \end{tabular}
    \caption{Prompt of \metricContext\label{tab:prompt_metric2}}
\end{table*}

\subsection{Prompts Used in Ablation}
\label{app:prompt_ablation}
See Table \ref{tab:prompt_fewshot} for more detail.

\begin{table*}[ht]
    \centering
    \begin{tabular}{|p{\textwidth}|}
    \hline
    {Do the following statements contradict each other? Answer ``Yes'' or ``No''.}\\\\
    
    Fact 1: John's lifestyle is strictly aligned with the teachings of his faith.\\
    Fact 2: John holds certain religious beliefs before his encounter with the entities.\\
    Answer: No\\\\
    
    Fact 1: The society in Europe was functioning normally without any widespread fear or despair.\\
    Fact 2: The populace of Europe is living in fear and despair due to the Black Death.\\
    Answer: Yes\\\\
    
    Fact 1: Emily was living a normal life without any chaos or fear related to supernatural experiences.\\
    Fact 2: The demon inside Emily had a certain level of control over her.\\
    Answer: Yes\\\\
    
    Fact 1: The selection process has started.\\
    Fact 2: The selection process continues to progress.\\
    Answer: No\\\\
    
    Fact 1: The footage contains information that can be analyzed.\\
    Fact 2: John has access to the footage from the camera.\\
    Answer: No\\\\
    
    Fact 1: The townsfolk are healthy and not infected with the mysterious virus.\\
    Fact 2: The infection is causing the townsfolk to behave strangely.\\
    Answer: Yes\\\\
    
    Fact 1: \textbf{\{fact1\}}\\
    Fact 2: \textbf{\{fact2\}}\\
    Answer: \\\\
    \hline
    \end{tabular}
    \caption{Prompt of fewshot learning in our ablation study.\label{tab:prompt_fewshot}}
\end{table*}
\section{Experiment Statistics}
\label{app:exp_stat}
We used 100 prompts to generate 90 Outlines from the writing prompt, with statistics in Table \ref{tab:outline_statistics}.
For \ours{} deployed on \modelLLaMA, we get 4589 positive pairs in total; for \ours{} deployed on \modelGPTF{}, we get 8883 positive pairs in total; details are shown in Table \ref{tab:facttrack_statistics}. For \baselineOutline{} on \modelGPTF{}, we get 408 positive pairs in total; for \baselinePairwise{} on \modelLLaMA, we subsample 10\% of the pairs and get 400 positive pairs, meaning if we run the whole population, we expect to get around 4,000 positive pairs for this baseline. 
Since we cannot easily change the threshold for our baseline, to make the comparison fairer, we subsample our method's predictions so that both methods flag a similar total number of contradictions.
To compare with \baselineOutline{} on \modelGPTF{}, which detected 298 positive pairs, we subsample by the \textit{Max fact pair NLI score} between event pairs for 300 samples. While comparing with \baselineOutline{} on \modelGPTT{} (829 positive pairs) and \baselinePairwise{} on \modelLLaMA{} (estimated 4000 positive pairs), we perform pure random sampling for 500 samples as a representative of the whole population (4589 positive pairs) of our method.

Tables \ref{tab:pairwise_distribution} and \ref{tab:outline_distribution} show the sample size and distribution of our evaluation set. Several data points failed during generation, and we simply dropped them.
By observing the distribution, we find that \ours{}'s detection capability is slightly inferior to \modelGPTF{} on clear contradictions (score 5), which may also be because we subsampled the top 300 instead of the top 298. However, on more nuanced or partial contradictions (scores 2, 3, and 4), \ours{} is significantly better than the two baselines. This observation intuitively confirms the strength of our approach.

Given the label-wise result, it is also possible to estimate a precision and recall value of the binary metrics. According to Tables \ref{tab:pairwise_distribution}, if we consider scores 4 and 5 as true positives and scores 1 to 3 as false positives, and assume there are 450 gold contradictions in all 90 outlines, our baseline of GPT-4 has $TP = 46$, $FP = 252$, $FN = 404$, with $P = 0.154$ and $R = 0.102$. \ours{} on GPT-4 has $TP = 123$, $FP = 177$, $FN = 327$, with $P = 0.410$ and $R = 0.273$, using the results in Table 13. However, this is only a rough estimate because the annotation of GPT-4 is not necessarily equivalent to the gold label, and the estimation of the total contradictions also requires precise annotation.

\label{app:experiment_stat}

\begin{table*}[ht]
    \centering
    \small
    \setlength\tabcolsep{2pt}
    %\resizebox{\columnwidth}{!}{
    \begin{tabular}{llcc}
    \toprule
    Statistics Name & & \multicolumn{2}{c}{Value} \\
    \midrule
    \# Outlines: & & \multicolumn{2}{c}{90} \\
    Events / Outline: & & \multicolumn{2}{c}{39} \\
    Words / Outline: & & \multicolumn{2}{c}{2490.5{\tiny $\pm$594.6}} \\
    Sentences / Outline: & & \multicolumn{2}{c}{85.68{\tiny $\pm$21.49}} \\
    Words / Sentence: & & \multicolumn{2}{c}{29.07} \\
    Unique Words / Outline: & & \multicolumn{2}{c}{382.51{\tiny $\pm$89.58}} \\
    \# Unique Words: & & \multicolumn{2}{c}{8766} \\
    \bottomrule
    \end{tabular}
    %}
    \caption{Statistics of outlines. Each outline contains three levels (3+9+27=39). Word and sentence counts were determined using NLTK for tokenization. The variance in word count across different outlines has been calculated. Unique words were identified by simple deduplication after tokenization. On average, each unique word appears 6.5 times within an outline and 25.6 times across all 90 outlines.}
    \label{tab:outline_statistics}
\end{table*}
\begin{table*}[ht]
    \centering
    \small
    \setlength\tabcolsep{2pt}
    \begin{tabular}{llcc}
    \toprule
    & & \multicolumn{1}{c}{\ours{} LLaMA2-7B} & \multicolumn{1}{c}{\ours{} GPT-4} \\
    \midrule
    \# Total Contradictions: & & 4559 & 8883 \\
    Positive Rate: & & 7\% & 14\% \\
    Avg. Contradictions per Outline: & & 50.66{\tiny $\pm$21.79} & 98.7{\tiny $\pm$34.93} \\
    Layer 1: & & 52 {\tiny 19\%} & 44 {\tiny 16\%} \\
    Layer 2: & & 313 {\tiny 10\%} & 623 {\tiny 20\%} \\
    Layer 3: & & 2229 {\tiny 7\%} & 4462 {\tiny 15\%} \\
    Layer 1 $\leftrightarrow$ Layer 2: & & 184 {\tiny 8\%} & 303 {\tiny 13\%} \\
    Layer 1 $\leftrightarrow$ Layer 3: & & 333 {\tiny 5\%} & 702 {\tiny 10\%} \\
    Layer 2 $\leftrightarrow$ Layer 3: & & 1448 {\tiny 7\%} & 2749 {\tiny 13\%} \\
    \bottomrule
    \end{tabular}
    \caption{Contradictions detected for \ours{} LLaMA2-7B and \ours{} GPT-4 by \ours{} with the settings in Appendix \ref{app:prompt_decomposition}. At the same NLI threshold, GPT-4 performs better at contradiction detection than LLaMA2-7B, capturing more potential contradictions due to fine-grained and accurate decomposition. We also visualize the frequency of contradictions occurring between different layers of the outline. Additionally, this provides insight into the higher-density contradictions that may occur in a higher-level outline.}
    \label{tab:facttrack_statistics}
\end{table*}

\begin{table*}[t]
\centering \tiny
\begin{tabular}{lccccccc}
\toprule
\text{Method} & N (Magnitude) & N (Sample Size) & \text{(1)} & \text{(2)} & \text{(3)} & \text{(4)} & \text{(5)} \\
\midrule
\baselineOutline{} (\modelGPTF{}) & 298 & 296 & 35.14\% & 34.12\% & 15.20\% & 9.46\% & 6.08\% \\
\ours{} (\modelGPTF{}, top 300) & 300 & 300 & 4.33\% & 29.67\% & 25.00\% & 30.33\% & 10.67\% \\
\ours{} (\modelLLaMA{}, top 300) & 300 & 300 & 36.00\% & 30.00\% & 15.00\% & 11.33\% & 7.67\% \\
\text{\ours{} (\modelLLaMA{}, random 500)} & 4589 & 500 & 44.60\% & {42.20\%} & 3.00\% & 5.40\% & 4.80\% \\
\baselineOutline{} (\modelGPTT{}) & 829 & 500 & 56.40\% & 36.40\% & 3.20\% & 3.20\% & 0.80\% \\
\text{\baselinePairwise{}(\modelLLaMA{})} & 4000 & 400 & 67.25\% & 26.25\% & 2.25\% & 2.50\% & 1.75\% \\
\text{Random} & 66690 & 408 & {68.87\%} & 25.49\% & 1.96\% & 2.21\% & 1.47\% \\
\bottomrule
\end{tabular}
\caption{Distribution of \metricPairwise{} for different methods. Magnitude refers to how many positive samples are detected by a given method across 90 outlines. Sample size is the number of randomly selected samples we allowed \modelGPTF{} to annotate.% \kevin{I changed the wording here, you can double check}
}
\label{tab:pairwise_distribution}
\end{table*}

\begin{table*}[t]
\centering \tiny
\begin{tabular}{lccccccc}
\toprule
\text{Method} & N (Magnitude)& N (Sample Size) & \text{(1)} & \text{(2)} & \text{(3)} & \text{(4)} & \text{(5)} \\
\midrule
\baselineOutline{} (\modelGPTF{}) & 298 & 298 & 34.90\% & 23.49\% & 14.77\% & 13.42\% & 13.42\% \\
\ours{} (\modelGPTF{}, top 300) & 300 & 299 & 21.07\% & 37.13\% & 14.72\% & 15.05\% & 12.04\% \\
\ours{} (\modelLLaMA{}, top 300) & 300 & 300 & 31.00\% & 22.00\% & 18.00\% & 17.67\% & 11.33\% \\
\text{\ours{} (\modelLLaMA{}, random 500)} & 4589 & 500 & 33.40\% & {44.40\%} & 10.20\% & 8.20\% & 3.80\% \\
\baselineOutline{} (\modelGPTT{}) & 829 & 500 & 48.40\% & 35.00\% & 9.20\% & 6.20\% & 1.20\% \\
\text{\baselinePairwise{} (\modelLLaMA{})} & 4000 & 400 & 55.25\% & 33.25\% & 5.25\% & 4.75\% & 1.50\% \\
\text{Random} & 66690 & 408 & 57.35\% & 30.39\% & 7.11\% & 3.19\% & 1.96\% \\
\bottomrule
\end{tabular}
\caption{Distribution of \metricContext{}. Magnitude refers to how many positive samples were detected by a given method out of 90 outlines. Sample size is the number of randomly selected samples we allowed \modelGPTF{} to annotate.}
\label{tab:outline_distribution}
\end{table*}

\section{Using \ours{} to Enhance Outline Generation}
\label{app:outline_rewrite}
We also implement algorithms to enhance outline generation by direct rewriting, rewriting conditional on facts (see Table \ref{tab:prompt_rewrite}), and rewriting conditional on events (see Table \ref{tab:prompt_fact_retrieve}). Qualitative inspection shows that this approach can improve factual consistency since it allows us to detect and solve problems at a high level of planning. Sometimes our method can make false positive predictions, and continuing rewrite, we apply one of the following two strategies: (1) rewrite the event until rewrite sampling time (by default 5 times) and then ignore it; (2)  resample the event until maximum sampling time (by default 10 times) and then rewrite the whole outline. Both strategies can help generate a high-quality outline.

\begin{table*}[ht]
    \centering
    \begin{tabular}{|p{\textwidth}|}
    \hline
    Below is a event point which contradicts one or more Existing Facts. Please rewrite the event point to align with all Existing Facts, while keeping as much of the original information as possible and maintaining a clear and concise description. \\
\\
event point: \textbf{\{curr\_event\}}\\
\\
Existing Facts:\\
    \begin{minipage}[t]{\linewidth}
        \begin{enumerate}
        \item \textbf{\{status(fact\_1)\}}, \textbf{\{fact\_1\}}
        \item \textbf{\{status(fact\_2)\}}, \textbf{\{fact\_2\}}
        \item ...\\
        \end{enumerate}
    \end{minipage}\\
    \hline
    \end{tabular}
    \caption{The prompt given to \modelGPTF{} to inject facts to  a given event\label{tab:prompt_rewrite}}
\end{table*}

\begin{table*}[ht]
    \centering
    \begin{tabular}{|p{\textwidth}|}
    \hline
        Below is a Current event point. Please rewrite it to make it more consistent with the given Existing event points, taking into account that the outline is structured as a tree. In this tree-like structure, individual points such as 1.1, 1.2, and 1.3 are child nodes of event point 1. Retain as much of the original content as possible, and maintain clarity and coherence.
\\
\\
Current event point \textbf{\{curr\_event\_idx\}}: \textbf{\{curr\_event\}}\\
\\
Existing event points:\\
    \begin{minipage}[t]{\linewidth}
        \begin{enumerate}
        \item event point \textbf{\{event\_1\_idx\}}: \textbf{\{event\_1\}}
        \item event point \textbf{\{event\_2\_idx\}}: \textbf{\{event\_2\}}
        \item ...\\
        \end{enumerate}
    \end{minipage}\\
    \hline
    \end{tabular}
    \caption{The prompt given to \modelGPTF{} to retrieve events to make a given event more consistent\label{tab:prompt_fact_retrieve}}
\end{table*}

% \section{Example of Decomposition}
% \label{tab:decomposition_example}

% \section{Examples of validity interval}
% \label{tab:valid_interval_example}

\section{More Examples}
\label{app:example}
We show five correct examples of contradictions we detected in a story outline in Table \ref{tab:example_1}, 
Table \ref{tab:example_2}, Table \ref{tab:example_4}, Table \ref{tab:example_5} and \ref{tab:example_6}. We can see that our approach performs detailed comparisons at a fine-grained fact level rather than simply checking for contextual continuity.

\begin{table*}[t]
\small
\begin{tabular}{p{0.95\linewidth}lllllll}
\toprule
\textbf{Event Pair}\\
\toprule
\texttt{\textbf{event 1.2.1 "Discovery of Unusual Side Effects":} Dr. Maria Rodriguez discovers that the building's energy field is causing unusual side effects in the townspeople, including vivid dreams and altered states of consciousness. \textit{At the beginning}, Dr. Rodriguez notices that some of the townspeople are reporting strange experiences after being near the building. \textit{At end}, Dr. Rodriguez conducts experiments to determine the cause of the side effects and discovers that they are related to the building's energy field.}

\texttt{\textbf{event 2: Main event - Unexpected Consequences of the Building's Energy Field.} At the beginning, Dr. Rodriguez discovers that the building's energy field is causing unusual side effects in the townspeople, including vivid dreams and altered states of consciousness. \textit{At end}, James confides in Sarah about his strange dreams, and she realizes that the building's energy field may be linked to the increasingly bizarre occurrences in the town.}\\
\toprule
\textbf{Atomic Fact Pairs Detected}\\
\toprule
\texttt{\textbf{Fact 1:} The energy field is altering the townspeople's brain activity, leading to vivid dreams and altered states of consciousness., \textbf{Fact 2:} The townspeople have been living near the building for several years without any issues., } \textcolor{blue}{\quad$P(\textrm{contradict})$: \textit{0.8462}}\\
\midrule
\texttt{\textbf{Fact 1:} The energy field is altering the townspeople's brain activity, leading to vivid dreams and altered states of consciousness., \textbf{Fact 2:} The energy field is not harmful to humans., } \textcolor{blue}{\quad$P(\textrm{contradict})$: \textit{0.7984}}\\
\midrule
\texttt{\textbf{Fact 1:} Dr. Rodriguez identifies the specific frequency of the energy field that is causing the side effects., \textbf{Fact 2:} The energy field is emitting a unique frequency that is not harmful to humans., } \textcolor{blue}{\quad$P(\textrm{contradict})$: \textit{0.7816}}\\
\midrule
\texttt{\textbf{Fact 1:} Dr. Rodriguez identifies the specific frequency of the energy field that is causing the side effects., \textbf{Fact 2:} The energy field is not harmful to humans.,} \textcolor{blue}{\quad$P(\textrm{contradict})$: \textit{0.2543}}\\
\midrule
\texttt{\textbf{Fact 1:} The townspeople are experiencing unusual side effects after being near the building., \textbf{Fact 2:} The townspeople have been living near the building for several years without any issues., } \textcolor{blue}{\quad$P(\textrm{contradict})$: \textit{0.8462}}\\
\bottomrule
\caption{Example 1. The contradiction occurs because in Event 1.2.1, unusual side effects have already been discovered and attributed to the building's energy field, but in Event 2, it is implied that the causal relationship between the building and the side effects has not yet been discovered.}
\label{tab:example_1}
% \vspace{-2em}
\end{tabular}
\end{table*}

\begin{table*}[t]
\small
\begin{tabular}{p{0.95\linewidth}lllllll}
\toprule
\textbf{Event Pair}\\
\toprule
\texttt{\textbf{event 2.3.1 The Final Challenge:} The contestant faces a final, climactic challenge that tests their skills and determination in a dramatic and intense way. \textit{At the beginning}, The contestant receives word of the final challenge and must prepare themselves mentally and physically. \textit{At end}, The contestant completes the final challenge and is declared the winner of the contest.}

\texttt{\textbf{event 2.3.2 The Mentor's Support:} The contestant receives guidance and support from a helpful character who provides valuable advice and encouragement throughout the final challenge. \textit{At the beginning}, The contestant encounters a particularly difficult part of the final challenge and seeks the mentor's help. \textit{At end}, The contestant successfully completes the final challenge with the mentor's support.}\\
\toprule
\textbf{Atomic Fact Pairs Detected}\\
\toprule
\texttt{\textbf{Fact 1:} The contestant has completed the final challenge and received the outcome of their performance., \textbf{Fact 2:} The contestant is facing a difficult part of the final challenge., } \textcolor{blue}{\quad$P(\textrm{contradict})$: \textit{0.7638}}\\
\midrule
\texttt{\textbf{Fact 1:} The contestant has completed the final challenge and received the outcome of their performance., \textbf{Fact 2:} The contestant has not yet completed the final challenge., } \textcolor{blue}{\quad$P(\textrm{contradict})$: \textit{0.9736}}\\
\midrule
\texttt{\textbf{Fact 1:} The contestant has achieved a significant milestone in their career or personal growth as a result of their performance in the challenge., \textbf{Fact 2:} The contestant has not yet completed the final challenge., } \textcolor{blue}{\quad$P(\textrm{contradict})$: \textit{0.7533}}\\
\bottomrule
\caption{Example 2. Although these two events were generated under one query of LLM, and there is still a factual contradiction. Event 2.3.1 indicates that the protagonist has completed the final challenge, but Event 2.3.2 suggests his mentor is helping him with the final challenge, creating a contradiction."}
\label{tab:example_2}
% \vspace{-2em}
\end{tabular}
\end{table*}

\begin{table*}[t]
\small
\begin{tabular}{p{0.95\linewidth}lllllll}
\toprule
\textbf{Event Pair}\\
\toprule
\texttt{\textbf{event 2.3:} Echo investigates the hidden message and uncovers a conspiracy involving Dr. Kim and the government. \textit{At the beginning}, Echo decodes the hidden message and begins to investigate its contents. \textit{At end}, Echo discovers a sinister event involving Dr. Kim and the government.}

\texttt{\textbf{event 3.2.2:} Echo investigates the purpose of the secret government project. \textit{At the beginning}, Echo finds a series of encrypted messages in the hidden folder. \textit{At end}, Echo decrypts the messages and learns about the government's involvement in Dr. Kim's work.}\\
\toprule
\textbf{Atomic Fact Pairs Detected}\\
\toprule
\texttt{\textbf{Fact 1:} Echo has uncovered a conspiracy involving Dr. Kim and the government., \textbf{Fact 2:} Echo has no knowledge of the government's involvement in Dr. Kim's work., } \textcolor{blue}{\quad$P(\textrm{contradict})$: \textit{0.9347}}\\
\bottomrule
\caption{Example 3. The contradiction arises because while Echo has already uncovered a conspiracy involving Dr. Kim and the government after event 2.3, but in event 3.2.2 there's an implication that Echo is unaware of the government's involvement in Dr. Kim's work.}
% \vspace{-2em}
\label{tab:example_4}
\end{tabular}
\end{table*}

\begin{table*}[t]
\small
\begin{tabular}{p{0.95\linewidth}lllllll}
\toprule
\textbf{Event Pair}\\
\toprule
\texttt{\textbf{event 2.3:} As they reminisce about their past, Marcus and Leon begin to see each other in a new light, and their animosity towards each other starts to fade. \textit{At the beginning}, Marcus is surprised by Leon's kindness and vulnerability, and begins to see him as a person, not just an enemy. \textit{At end}, Leon reciprocates, and they both feel a sense of camaraderie and understanding.}

\texttt{\textbf{event 3:} As Marcus and Leon approach death, they begin to question the reasons behind their conflict and the true cost of war. \textit{At the beginning}, Marcus wonders if there was another way to resolve the conflict without resorting to violence. \textit{At end}, Leon reflects on the sacrifices they have made and the lives they have lost, and hopes that their deaths will not be in vain.}\\
\toprule
\textbf{Atomic Fact Pairs Detected}\\
\toprule
\texttt{\textbf{post-fact 1:} Leon reciprocates Marcus's new perspective on him., \textbf{pre-fact 1:} Marcus and Leon are in conflict with each other., \textcolor{blue}{\quad$P(\textrm{contradict})$: \textit{0.7834}}}\\
\midrule
\texttt{\textbf{post-fact 2:} Marcus no longer views Leon as an enemy., \textbf{pre-fact 2:} Marcus and Leon are mortal enemies., \textcolor{blue}{\quad$P(\textrm{contradict})$: \textit{0.9566}}}\\
\bottomrule
\caption{Example 4. The contradiction arises in event 2.3, where Marcus and Leon have already reconciled, compared to event 3, where they just ``begin to question the reasons behind their conflict,'' indicating they are still in conflict.}
\vspace{-2.5em}
\label{tab:example_5}
% \vspace{-5em}
\end{tabular}
\end{table*}

\begin{table*}[t]
%\vspace{-3em}
\small
\begin{tabular}{p{0.95\linewidth}lllllll}
\toprule
\textbf{Event Pair}\\
\toprule
\texttt{\textbf{event 1.3.3:} As the group learns more about their shared destiny, they begin to uncover secrets about their pasts that have been hidden from them. \textit{At the beginning}, The group discovers that their shared destiny is connected to a larger conspiracy involving a powerful organization. \textit{At end}, The group learns the truth about their pasts and the reason they have been brought together, and must decide how to use their newfound knowledge to change their lives and the world around them.}

\texttt{\textbf{event 3.2:} As the group of strangers continues on their journey, they begin to uncover hidden secrets about their pasts and the mysterious force that brought them together. They must work together to unravel the truth before it's too late. \textit{At the beginning}, The group discovers a cryptic message that seems to point to a sinister event involving their shared destiny. \textit{At end}, The group uncovers a shocking truth about their pasts and the true nature of the force that brought them together..}\\
\toprule
\textbf{Atomic Fact Pairs Detected as Contradict}\\
\toprule
\texttt{\textbf{post-fact 1:} The group learns secrets about their pasts that have been hidden from them., \textbf{pre-fact 1:} They have no memory of their past or how they were brought together., } \textcolor{blue}{\quad$P(\textrm{contradict})$: \textit{0.8495}}\\
\midrule
\texttt{\textbf{post-fact 2:} The group learns secrets about their pasts that have been hidden from them., \textbf{pre-fact 2:} They are unaware of any hidden secrets about their pasts, } \textcolor{blue}{\quad$P(\textrm{contradict})$: \textit{0.9822}}\\
\midrule
\texttt{\textbf{post-fact 3:} The group begins to understand the reason they have been brought together \textbf{pre-fact 3:} They have no memory of their past or how they were brought together. } \textcolor{blue}{\quad$P(\textrm{contradict})$: \textit{0.4507}}\\
\midrule
\texttt{\textbf{post-fact 4:} The group must work together to uncover the truth about their pasts and their destiny., \textbf{pre-fact 4:} They are unaware of any hidden secrets about their pasts, } \textcolor{blue}{\quad$P(\textrm{contradict})$: \textit{0.6361}}\\
\bottomrule
\caption{Example 5. 
The contradiction arises because in event 1.3.3, the group already gains knowledge of hidden secrets that are the reason they have been brought together, which contradicts the indication that before event 3.2, they are unaware about those secrets.}%\kevin{this explanation is a bit confusing. why is it a contradiction to discover hidden secrets just because you previously didn't know about them?}}
\vspace{-2.5em}
\label{tab:example_6}
\end{tabular}
\end{table*}

% \section{Examples of Disagreement of \modelGPTF{} and Our Method}
% \label{tab:valid_interval_example}

\begin{table*}[t]
\small
\begin{tabular}{p{0.95\linewidth}lllllll}
\toprule
\textbf{Event Pair}\\
\toprule
\texttt{\textbf{event 1.3: Main event - Investigating the Building's Origins.} At the beginning, Agent Thompson uncovers evidence that the building in Sarah's town is not an isolated incident, and that there are similar structures appearing all over the world. \textit{At end}, Agent Thompson realizes that the buildings are not of this world, and that they are connected to an ancient civilization with advanced technology.}

\texttt{\textbf{event 3.1.1 "Uncovering the Hidden Message":} Agent Thompson investigates the mysterious buildings globally and discovers a hidden message in one of them that leads him to believe they are not of this world. \textit{At the beginning}, Agent Thompson finds a hidden compartment in one of the buildings that contains a cryptic message. \textit{At end}, Agent Thompson deciphers the message and realizes it points to an ancient civilization with advanced technology.}\\
\toprule
\textbf{Atomic Fact Pairs Detected}\\
\toprule
\texttt{\textbf{Fact 1:} The buildings are connected to each other, forming a vast network of interconnected structures., \textbf{Fact 2:} The buildings are globally distributed, with no discernible pattern or connection between them., } \textcolor{blue}{\quad$P(\textrm{contradict})$: \textit{0.9450}}\\
\bottomrule
\caption{Failure case 1 for \ours{}. The problem here is ``Decompose is not atomic enough''. It is ambiguous to say ``connected to each other'' in reality or on a metaphysical level. }
% \vspace{-2em}
\label{tab:badcase1}
\end{tabular}
\end{table*}

\begin{table*}[t]
\small
\begin{tabular}{p{0.95\linewidth}lllllll}
\toprule
\textbf{Event Pair}\\
\toprule
\texttt{\textbf{event: 1.3.1 - The Owner and Whiskers Share Memories of Their Past Adventures:} The owner and Whiskers spend time reminiscing about their past adventures and experiences together. \textit{At the beginning}, The owner and Whiskers sit together, looking through old photos and mementos from their time together. \textit{At end}, The owner and Whiskers laugh and smile as they remember their favorite memories with each other..}

\texttt{\textbf{event 2.2.3 Goodbye Moment:} The owner says goodbye to Whiskers. \textit{At the beginning}, The owner approaches Whiskers, looking sad and tearful. \textit{At end}, The owner pets Whiskers one last time, whispering words of love and appreciation..}\\
\toprule
\textbf{Atomic Fact Pairs Detected}\\
\toprule
\texttt{\textbf{Fact 1:} The owner and Whiskers are smiling and laughing as they remember their favorite memories with each other., \textbf{Fact 2:} The owner is sad and tearful., } \textcolor{blue}{\quad$P(\textrm{contradict})$: \textit{0.8451}}\\
\bottomrule
\caption{Failure case 2 for \ours{}. The problem here is ``Facts are not updated in the timeline.''. The owner's mood changes can be found in the outline, but they don't seem to be well tracked and updated in \ours{}.}
% \vspace{-2em}
\label{tab:badcase2}
\end{tabular}
\end{table*}

\begin{table*}[t]
\small
\begin{tabular}{p{0.95\linewidth}lllllll}
\toprule
\textbf{Event Pair}\\
\toprule
\texttt{\textbf{event: 2.1.3:} Jack's Search for Shelter and Supplies. \textit{At the beginning}, Jack and Sarah search for a safe haven, but they soon realize it's not as secure as they thought. \textit{At end}, Jack and Sarah must find a new place to hide and regroup, while also trying to uncover the truth about the explosion and those responsible.}

\texttt{\textbf{event: 2.3:} Finding Shelter. \textit{At the beginning}, Jack and Sarah continue their journey, searching for a safe haven, but they soon realize it's not as secure as they thought. \textit{At end}, Jack and Sarah must find a new place to hide and regroup, while also trying to uncover the truth about the explosion and those responsible.}\\
\toprule
\textbf{Atomic Fact Pairs Detected}\\
\toprule
\texttt{\textbf{Fact 1:} Jack and Sarah have found a new place to hide and regroup, but it is not as secure as they thought., \textbf{Fact 2:} They have been searching for a while, but have not found a suitable place yet., } \textcolor{blue}{\quad$P(\textrm{contradict})$: \textit{0.5394}}\\
\bottomrule
\caption{Failure case 3 for \ours{}. The problem here is ``Contradiction detection makes a mistake.''. }
% \vspace{-2em}
\label{tab:badcase3}
\end{tabular}
\end{table*}

\section{Error Analysis}
\label{app:badcase}
Although our method significantly outperforms the baseline, it still fails in certain cases. These failures include:
\begin{enumerate}
\item Decomposition is not sufficiently atomic;
\item Mistakes in contradiction detection;
\item Facts are not updated in the timeline.
\end{enumerate}
Table \ref{tab:badcase1}, \ref{tab:badcase2} and \ref{tab:badcase1} illustrate three examples of such failures.

\section{Usage of AI Assistants}
During the development process, we drafted the code framework and data structures by ourselves initially, using Copilot to assist in enhancing development efficiency. Additionally, in the writing phase, we only employed ChatGPT for proofreading purposes, such as correcting grammatical errors and properly formatting tables.

\section{Model Size and Computation Budget}
We utilized the LLaMA2-7B-Chat model, a 7 billion parameter generative language model. Additionally, the NLI and retrieval models used were both under 1 billion parameters. The total computational cost for all experiments did not exceed 100 GPU hours on an A6000, and the cost of API of GPT-series was kept under \$500 USD.

\end{document}